\documentclass{article}

\usepackage[final]{corl_2019} % Uncomment for the camera-ready ``final'' version

\usepackage[utf8]{inputenc} % allow utf-8 input
\usepackage[T1]{fontenc}    % use 8-bit T1 fonts
\usepackage{hyperref}       % hyperlinks
\usepackage{url}            % simple URL typesetting
\usepackage{booktabs}       % professional-quality tables
\usepackage{amsfonts}       % blackboard math symbols
\usepackage{nicefrac}       % compact symbols for 1/2, etc.
\usepackage{microtype}      % microtypography

\usepackage{times}
\usepackage{epsfig}
\usepackage{graphicx}
\usepackage{amsmath}
\usepackage{amssymb}
\usepackage{gensymb}

\usepackage{units}
\usepackage{subcaption}
\usepackage{multirow}
\usepackage{bm}
\usepackage{mydefs}
\usepackage{bbm}
\usepackage{pifont}
\usepackage{xspace}

\makeatletter
\DeclareRobustCommand\onedot{\futurelet\@let@token\@onedot}
\def\@onedot{\ifx\@let@token.\else.\null\fi\xspace}

\def\eg{\emph{e.g}\onedot} 
\def\ie{\emph{i.e}\onedot} 
 
\def\etc{\emph{etc}\onedot} 
 
\def\etal{\emph{et al}\onedot}
\makeatother

\newcommand{\out}[1]{}

\newcommand{\multiflow}{MultiPath\xspace}

\newcommand{\supsecref}[1]{\secref{#1}}

\title{MultiPath: Multiple Probabilistic Anchor Trajectory Hypotheses for Behavior Prediction}

\newcommand*\samethanks[1][\value{footnote}]{\footnotemark[#1]}

\author{
  Yuning Chai\thanks{Equal contribution} \qquad Benjamin Sapp\samethanks \qquad Mayank Bansal \qquad Dragomir Anguelov \\
  \\
  Waymo LLC \\
  \texttt{\{chaiy,bensapp\}@waymo.com} \\
}

\begin{document}
\maketitle

\begin{abstract}
Predicting human behavior is a difficult and crucial task required for motion planning.  It is challenging in large part due to the highly uncertain and multi-modal set of possible outcomes in real-world domains such as autonomous driving.  Beyond single MAP trajectory prediction~\cite{Luo18, Casas18}, obtaining an accurate probability distribution of the future is an area of active interest~\cite{Lee17,Rhinehart18}. We present \multiflow, which leverages a fixed set of future state-sequence anchors that correspond to modes of the trajectory distribution.  At inference, our model predicts a  discrete distribution over the anchors and, for each anchor, regresses offsets from anchor waypoints along with uncertainties, yielding a Gaussian mixture at each time step. Our model is efficient, requiring only one forward inference pass to obtain multi-modal future distributions, and the output is parametric, allowing compact communication and analytical probabilistic queries. We show on several datasets that our model achieves more accurate predictions, and compared to sampling baselines, does so with an order of magnitude fewer trajectories.
 
 \out{Recent approaches either predict single trajectories~\cite{Luo18,Casas18}, or use a generative model to obtain multiple trajectory samples~\cite{Lee17,Rhinehart18}.  While flexible, sample-generating models have several issues: trajectory sample likelihoods are unknown, it is not obvious how many samples are sufficient to guarantee that important safety-critical modes are covered and finally, the quality of generated samples can be suboptimal in highly multi-modal real-world situations. 

In contrast, we present \multiflow, which leverages a fixed set of future state-sequence anchors that correspond to modes of the trajectory distribution.  At inference, our model predicts a  distribution over the anchors and regresses trajectories and associated uncertainties within each mode, yielding a Gaussian mixture at each time step. Our model is efficient, requiring only one forward inference pass to obtain multi-modal future distributions, and the output is parametric, allowing compact communication and analytical probabilistic queries. We show on several datasets that our model achieves more accurate predictions, and compared to sampling baselines, does so with an order of magnitude fewer trajectories.
}

\end{abstract}
\section{Introduction}
\label{sec:intro}

\begin{figure}[tb]
  \centering
\includegraphics[width=0.9\textwidth]{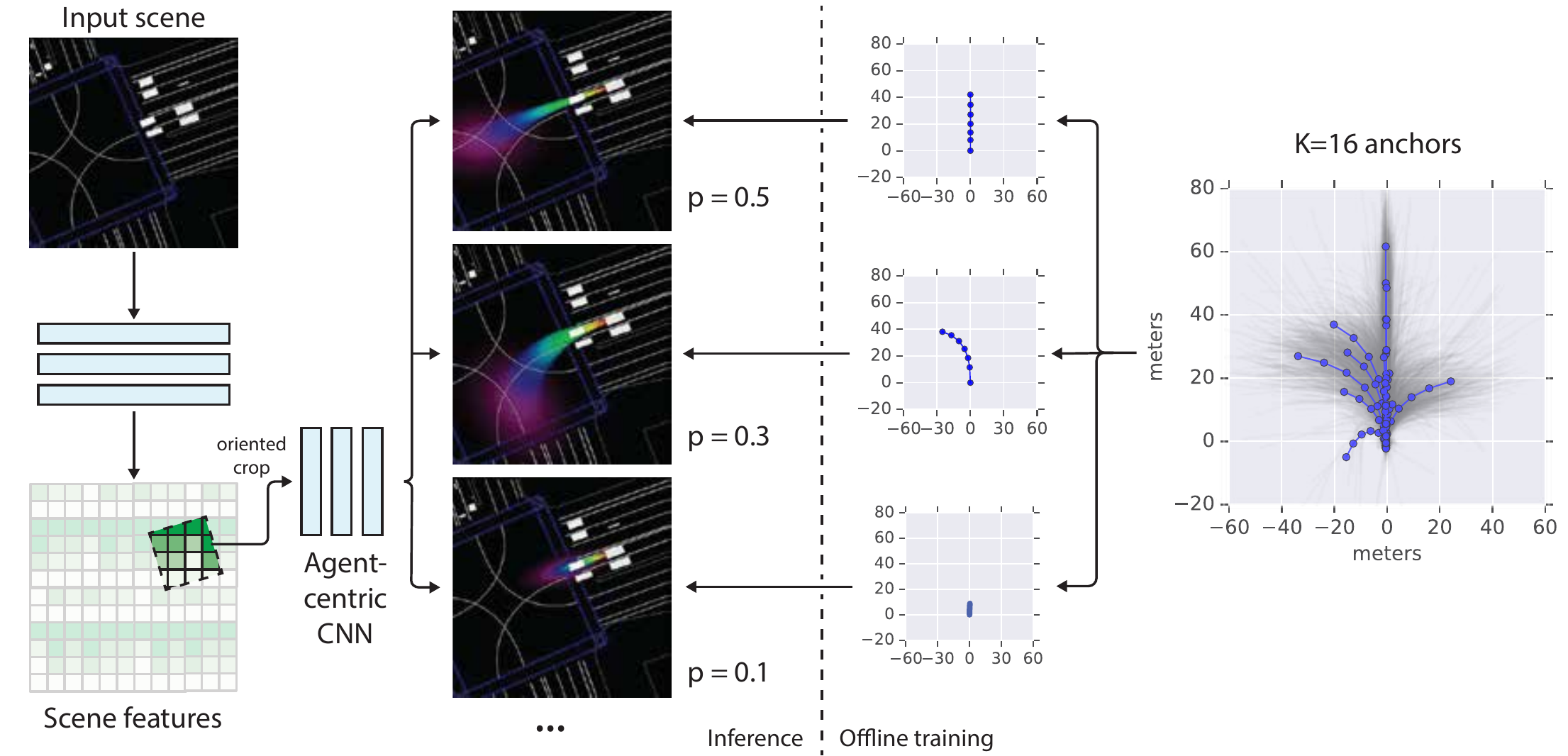}
\caption{\small
 \multiflow estimates the distribution over future trajectories per agent in a scene, as follows: 1) Based on a top-down scene representation, the Scene CNN extracts mid-level features that encode the state of individual agents and their interactions. 2) For each agent in the scene, we crop an agent-centric view of the mid-level feature representation and predict the probabilities over the fixed set of $K$ predefined anchor trajectories. 3) For each anchor, the model regresses offsets from the anchor states and uncertainty distributions for each future time step.
 \vspace{-0.8cm}}
\label{fig:teaser}
\end{figure}

We focus on the problem of predicting future agent states, which is a crucial task for robot planning in real-world environments.  We are particularly interested in addressing this problem for self-driving vehicles, an application with a potentially enormous societal impact. Importantly, predicting the future of other agents in this domain is vital for safe, comfortable and efficient operation.  For example, it is important to know whether to yield to a vehicle if they are going to cut in front of our robot or when would be the best time to merge into traffic. Such future prediction requires an understanding of the static and dynamic world context: road semantics (\eg, lane connectivity, stop lines), traffic light information, and past observations of other agents, as depicted in Fig.~\ref{fig:teaser}.

A fundamental aspect of future state prediction is that it is inherently {\em stochastic}, as agents cannot know each other's motivations.  When driving, we can never really be sure what other drivers will do next, and it is important to consider multiple outcomes and their likelihoods.  

We seek a model of the future that can provide both (1) a weighted, parsimonious set of discrete trajectories that covers the space of likely outcomes and (2) a closed-form evaluation of the likelihood of any trajectory.  These two attributes enable efficient reasoning in crucial planning use-cases, for example, human-like reactions to discrete trajectory hypotheses (\eg, yielding, following), and probabilistic queries such as the expected risk of collision in a space-time region.

Both of these attributes present modeling challenges.  Models which try to achieve diversity and coverage often suffer from mode collapse during training~\cite{Rhinehart18, Bishop06, Hong19}, while tractable probabilistic inference is difficult due to the space of possible trajectories growing exponentially over time.

Our MultiPath model addresses these issues with a key insight: it employs a fixed set of {\em trajectory anchors} as the basis of our modeling. This lets us factor stochastic uncertainty hierarchically: First, {\em intent uncertainty} captures the uncertainty of {\em what} an agent intends to do and is encoded as a distribution over the set of anchor trajectories. Second, given an intent, {\em control uncertainty} represents our uncertainty over {\em how} they might achieve it. We assume control uncertainty is normally distributed at each future time step~\cite{Thrun05}, parameterized such that the mean corresponds to a context-specific offset from the anchor state, with the associated covariance capturing the unimodal aleatoric uncertainty~\cite{Kendall17}.  Fig.~\ref{fig:teaser} illustrates a typical scenario where there are 3 likely intents given the scene context, with control mean offset refinements respecting the road geometry, and control uncertainty intuitively growing over time.

Our trajectory anchors are modes found in our training data in state-sequence space via unsupervised learning.  These anchors provide templates for coarse-granularity futures for an agent and might correspond to semantic concepts like ``change lanes'', or ``slow down'' (although to be clear, we don't use any semantic concepts in our modeling). 

Our complete model predicts a Gaussian mixture model (GMM) at each time step, with the mixture weights (intent distribution) fixed over time. Given such a parametric distribution model, we can directly evaluate the likelihood of any future trajectory and also have a simple way to obtain a compact, diverse weighted set of trajectory samples: the MAP sample from each anchor-intent.

\out{
In contrast to recent work predicting non-parametric occupancy grid distributions~\cite{Bansal19, Hong19}, it is a more compact description of the future by orders of magnitude.
}

Our model contrasts with popular past approaches which either provide only a single MAP trajectory~\cite{Luo18, Casas18, Sadeghian18, Pellegrini09, Bansal19} or an unweighted set of samples via a generative model~\cite{Lee17,  Rhinehart18, Hong19,  Ivanovic18, Bhattacharyya18BestOfMany, Kitani12, Ma17}.
There are a number of downsides to sample-based methods when it comes to real-world applications such as self-driving vehicles: (1) non-determinism in a safety critical system, (2) a poor handle on approximation error (e.g,. “how many samples must I draw to know the chance the pedestrian will jaywalk?”), (3) no easy way to perform probabilistic inference for relevant queries, such as computing expectations over a spacetime region.

We demonstrate empirically that our model emits distributions which predict the observed outcomes better on synthetic and real-world prediction datasets: we achieve higher likelihood than a model which emits unimodal parametric distributions, showing the importance of multiple anchors in real-world data. We also compare to sampling-based methods by using our weighted set of MAP trajectories per anchor, which describe the future better with far fewer samples on sample-set metrics.
\out{

Notably, a large amount of previous work has modeled a {\em deterministic} future, providing only a single best guess at where agents might be~\cite{Luo18, Casas18, Sadeghian18, Pellegrini09, Bansal19}, which is of limited practical use.

One way to capture the inherent stochasticity of the future is via a set of trajectory samples, which should cover the space of possible outcomes well with as few samples as possible.  A common way to obtain such samples is from rolling out 1-time-step stochastic policies~\cite{Rhinehart18, Ma17, Kitani12}, or stochastic latent-space random sampling models such as Conditional Variational Auto-Encoders (CVAEs)~\cite{Lee17, Hong19, Bhattacharyya18BestOfMany, Ivanovic18}.
}
\section{Related work} \label{sec:relwork}
We broadly categorize previous approaches to predicting future trajectory distributions into two classes of models: deterministic and stochastic. 
Deterministic models predict a single most-likely trajectory per agent, usually via supervised regression~\cite{Luo18, Casas18, Sadeghian18, Pellegrini09, Bansal19, Helbing95}.
\out{
Deterministic models predict a single most-likely trajectory for an agent. In the real-world driving domain, Fast and Furious~\cite{Luo18} and IntentNet~\cite{Casas18} regress single short-term future trajectories jointly with a model which performs detection and tracking. ChauffeurNet~\cite{Bansal19} applies a variety of domain-specific losses to improve single trajectory prediction for motion planning. In pedestrian modeling, the foundational Social Forces model~\cite{Helbing95} posed the problem as trajectory optimization through a hand-designed cost volume, and~\cite{Pellegrini09} extends this by learning the parameters of the cost volume. SocialLSTM~\cite{Sadeghian18} conditions an agent's next state on pooled neighbor internal states to explicitly model interactions.
}

Stochastic models incorporate random sampling during training and inference to capture future non-determinism. The seminal motion forecasting work of Kitani \etal~\cite{Kitani12} cast this as a Markov decision process and learns a 1-step policy, as does follow on work focusing on egocentric video and pedestrians~\cite{Ma17, Rhinehart17}. To encourage sample diversity and coverage, R2P2~\cite{Rhinehart18} proposes a symmetric KL loss between the predicted and data distributions. Several works explore the use of conditional variational autoencoders (CVAEs) and GANs to generate samples~\cite{Lee17, Hong19,Bhattacharyya18BestOfMany, Rhinehart19, Gupta19}. One drawback of such non-deterministic approaches is that they can make reproducing and analyzing results in a larger system difficult. 

Like us, a few previous works directly model probability distributions, either parametric~\cite{Hong19, Ivanovic18, Cui19} or in the form of probabilistic state-space occupancy grids (POGs)~\cite{Hong19, Bansal19}. While extremely flexible, POGs require state-space-dense storage to describe the distribution rather than just a few parameters, and it's not obvious how best to extract trajectory samples from POG space-time volumes. 

\out{
Finally, like our model, a few previous works directly model probability distributions. Cui \etal~\cite{Cui19} predict a discrete distribution over $K$ future trajectories (\ie, a weighted combination of Dirac delta functions, with probability zero almost everywhere), not a distribution over state space). Ivanovic \etal~\cite{Ivanovic18} focus specifically on predicting a sequence of GMMs, conditioned on a latent sampled variable (so a ``CVAE-GMM''). Relatedly,~\cite{Hong19} samples a latent discrete mode and predicts Gaussian uncertainty conditioned on that.  As an alternative, they also explore non-parametric distributions in the form of probabilistic state-space occupancy grids (POGs). While extremely flexible, POGs require state-space-dense storage to describe the distribution rather than just a few parameters, and it's not obvious how best to extract trajectory samples from POG space-time volumes. ChauffeurNet~\cite{Bansal19} models other agents with POGs in its multi-task network, but agents are not individually modeled.
}

Our method is influenced heavily by the concept of predefined anchors, which have a rich history in machine learning applications to handle multi-modal problems, starting with classic semi-parametric methods such as locally-weighted logistic regression, radial basis SVM and Gaussian Mixture Models~\cite{Bishop06}.  In the computer vision literature, they have been used effectively for detection~\cite{Erhan14} and human-pose estimation~\cite{Yang12}.  Like ours, these effective approaches predict the likelihood of anchors and also predict continuous refinements of state conditioned on these anchors (\eg box corners, joint locations or vehicle positions).

\section{Method}
\label{sec:method}

% \begin{figure*}[!tbp]
% \centering
% \includegraphics[width=0.7\textwidth]{images/arch.pdf}
% \caption{
% \TODO{?}}
% \label{fig:arch}
% \end{figure*}

Given observations $\bmx$ in the form of past trajectories of all agents in a scene and possibly additional contextual information (\eg, lane semantics, traffic light states)
, \multiflow seeks to provide (1) a parametric distribution over future trajectories $\bms$: $p(\bms | \bmx)$, and (2) a compact weighted set of explicit trajectories which summarizes this distribution well.

Let $t$ denote a discrete time step, and let $s_t$ denote the state of an agent at time $t$, the future trajectory $\bms = [s_1, \ldots, s_T]$ is a sequence of states from $t=1$ to a fixed time horizon $T$. We also refer to a state in a trajectory as a {\em waypoint}.

We factorize the notion of uncertainty into independent quantities. {\em Intent uncertainty} models uncertainty about the agents' latent coarse-scale intent or desired goal. For example, in a driving context, uncertainty about which lane the agent is attempting to reach. Conditioned on intent, there is still {\em control uncertainty}, which describes the uncertainty over the sequence of states the agent will follow to satisfy its intent.  Both intent and control uncertainty depend on the past observations of static and dynamic world context $\bmx$.

We model a discrete set of {\em intents} as a set of $K$ anchor trajectories $\setA=\{\bma^k\}_{k=1}^K$, where each anchor trajectory is a sequence of states: $\bma^k = [a^k_1, \ldots, a^k_T]$, assumed given for now.  We model uncertainty over this discrete set of intents with a softmax distribution: $ \pi(\bma^k | \bmx) = \frac{\exp{f_k(\bmx)}}{ \sum_i \exp{f_i(\bmx))}}$,
where $f_k(\bmx): \Real^{d(\bmx)} \mapsto \Real$ is the output of a deep neural network. 

We make the simplifying assumption that uncertainty is unimodal given intent, and model {\em control uncertainty} as a Gaussian distribution dependent on each waypoint state of an anchor trajectory:
\begin{align} 
\phi(s^k_t | \bma^k, \bmx) = \mathcal{N}(s^k_t | a^k_t + \mu^k_t(\bmx), \Sigma^k_t(\bmx))
\end{align}
The Gaussian parameters $\mu^k_t$ and $\Sigma^k_t$ are directly predicted by our model as a function of $\bmx$ for each time-step of each anchor trajectory $\bma^k_t$.  Note in the Gaussian distribution mean, $a^k_t + \mu^k_t$, the $\mu^k_t$ represents a scene-specific offset from the anchor state $a^k_t$; it can be thought of as modeling a scene-specific residual or error term on top of the prior anchor distribution. This allows the model to refine the static anchor trajectories to the current context, with variations coming from, \eg specific road geometry, traffic light state, or interactions with other agents.

The time-step distributions are assumed to be conditionally independent given an anchor, \ie, we write $\phi(s_t|\cdot)$ instead of $\phi(s_t|\cdot, s_{1:t-1})$. This modeling assumption allows us to predict for all time steps jointly with a single inference pass, making our model simple to train and efficient to evaluate. If desired, it is straightforward to add a conditional next-time-step dependency to our model, using a recurrent structure (RNN).

To obtain a distribution over the entire state space, we marginalize over agent intent:
\begin{align}
\label{eq:multiflow-likelihood}
    p(\mathbf{s} | \bmx) =  \sum_{k=1}^K \pi(\bma^k | \bmx) \prod_{t=1}^T  \phi(s_t | \bma^k, \bmx)
\end{align}
Note that this yields a Gaussian Mixture Model distribution, with mixture weights fixed over all time steps. This is a natural choice to model both types of uncertainty: it has rich representational power, a closed-form partition function, and is also compact.  It is easy to evaluate this distribution on a discretely sampled grid to obtain a probabilistic occupancy grid, more cheaply and with fewer parameters than a native occupancy grid formulation~\cite{Hong19, Bansal19}. 

\PAR{Obtaining anchor trajectories.}
% \label{sec:obtaining-anchors}
 Our distribution is parameterized by anchor trajectories $\setA$. As noted by \cite{Hong19, Bishop06}, directly learning a mixture suffers from issues of mode collapse. As is common practice in other domains such as object detection~\cite{Liu16} and human pose estimation~\cite{Yang12}, we estimate our anchors a-priori before fixing them to learn the rest of our parameters.
% From first principles, we wish to obtain a set of anchor trajectories which captures the training sample distribution of trajectories well.  If we consider the distribution induced by a kernel density estimator over the set of anchor trajectories, we can formalize anchor learning via estimating modes of a Gaussian Mixture Model via Expectation-Maximization. 
In practice, we used the k-means algorithm as a simple approximation to obtain $\setA$ with the following squared distance between trajectories: $d(\bmu, \bmv) = \sum_t^T ||M_u \bmu_t - M_v \bmv_t||_2^2$, where $M_u,M_v$ are affine transformation matrices which put trajectories into a canonical rotation- and translation-invariant agent-centric coordinate frame. In Sec.~\ref{sec:experiments}, on some datasets, k-means leads to highly redundant clusters due to prior distributions that are heavily skewed to a few common modes. To address this, we employ a simpler approach to obtain $\setA$ by uniformly sampling trajectory space.

% \subsection{Learning}
% \label{sec:learning}
\PAR{Learning.}
We train our model via {\em imitation learning} by fitting our parameters to maximize the log-likelihood of recorded driving trajectories. 
Let our data be of the form $\{(\bmx^m,\mathbf{\hat{s}}^m)\}_{m=1}^M$. We learn to predict distribution parameters $\pi(\bma^k|\bmx)$, $\mu(\bmx)^k_t$ and $\Sigma(\bmx)^k_t$ as outputs of a deep neural network parameterized by weights $\theta$ with the following negative log-likelihood loss built upon Equation~\ref{eq:multiflow-likelihood}:
\begin{align}
    \ell(\theta) =  - \sum_{m=1}^M \sum_{k=1}^K \mathbbm{1}(k=\hat{k}^m) \Bigl[ \log \pi(\bma^k |\bmx^m; \theta) +  \sum_{t=1}^T  \log \mathcal{N}(s^k_t | a^k_t + \mu^k_t, \Sigma^k_t; \bmx^m; \theta)\Bigr].
\end{align}
This is a time-sequence extension of standard GMM likelihood fitting~\cite{Bishop06}. The notation $\mathbbm{1}(\cdot)$ is the indicator function, and $\hat{k}^m$ is the index of the anchor most closely matching the groundtruth trajectory $\mathbf{\hat{s}}^m$, measured as $\ltwo$-norm distance in state-sequence space.  This hard-assignment of groundtruth anchors sidesteps the intractability of direct GMM likelihood fitting, avoids resorting to an expectation-maximization procedure, and gives practitioners control over the design of the anchors as they wish (see our choice below).  One could also employ a soft-assignment to anchors (\eg, proportional to the distance of the anchor to the groundtruth trajectory), just as easily.

\PAR{Inferring a diverse weighted set of test-time trajectories.}
% \label{sec:sample}
Our model allows us to eschew standard sampling techniques at test time, and obtain a weighted set of $K$ trajectories without any additional computation:  we take the MAP trajectory estimates from each of our $K$ anchor modes, and consider the distribution over anchors $\pi(\bma_k|\bmx)$ the sample weights (\ie, importance sampling).  When metrics and applications call for a set of top $\kappa < K$ trajectories for evaluation, we return the top $\kappa$ according to these sample weights.

\out{
\subsection{Sampling weighted trajectories}

To generate a discrete set of weighted trajectories, \ie, trajectories with scores that add up to one, we evaluate all $K$ clusters, yielding $K$ distinct trajectories $\setS$ in the form of $\bms^k=\bma^k + \mu(\bmx)^k$. Their likelihoods are given by the softmax functions $\pi(\bma^k|\bmx)$. Due to our choice of making the estimation of $\mu_t$ and $\Sigma_t$ conditionally independent in time, this model does not allow additional sampling. However, we usually choose a large enough $K$ so that we always end up with more trajectories than requested. If the model is required to return fewer trajectories than $K$, which is typically the case for most metrics discussed in the experiments, we simply sort the $K$ trajectories by their scores $\pi^k$. 
}

% \subsection{Probabilistic occupancy map prediction}

% Another popular representation for the behavior prediction task in autonomous driving is the probabilistic occupancy map that encodes the probability of a spatial location $i$ being occupied at a future time step $t$. This representation is powerful yet concise because it contains all the information needed for the motion planner to drive the autonomous agent. A \multiflow model trained with the MLE loss predicts a set of weighted trajectory with uncertainty per agent, which we can easily re-interpret as the probabilistic occupancy map by marginalizing over all trajectories and all agents. 

\PAR{Input representation.}% \label{sec:architecture}
We follow other recent approaches~\cite{Casas18, Hong19, Bansal19} and represent a history of dynamic and static scene context as a 3-dimensional array of data rendered from a top-down orthographic perspective. The first two dimensions represent spatial locations in the top-down image. The channels in the depth dimension hold static and time-varying (dynamic) content of a fixed number of previous time steps. Agent observations are rendered as orientated bounding box binary images, one channel for each time step. Other dynamic context such as traffic light state and static context of the road (lane connectivity and type, stop lines, speed limit, \etc) form additional channels. See \secref{sec:experiments} for further details, as the input content differs from dataset to dataset. An important benefit of using such a top-down representation is the simplicity of representing contextual information like the agents' spatial relationships to each other and semantic road information. In \supsecref{sec:roadgraph}, we empirically highlight its benefit towards behavior prediction.

\PAR{Neural network details.}
As shown in \figref{fig:teaser}, we designed a jointly-trained, two-stage architecture that first extracts a feature representation for the whole scene and then attends to each agent in the scene to make agent-specific trajectory predictions. 

The first stage is fully convolutional to preserve spatial structure; it takes the 3D input representation described above and outputs a 3D feature map of the entire top-down scene. We opt to use ResNet-based architectures~\cite{He16} for this scene-level feature extractor. We employ depth-wise thinned-out networks for all experiments, and a different number of residual layers depending on the dataset. See \supsecref{sec:backbone} for a speed-accuracy analysis of different ResNet setups.

The second phase extracts patches of size $11\times11$ centered on agents locations in this feature map. To be orientation invariant, the extracted features are also rotated to an agent-centric coordinate system via a differentiable bilinear warping. The efficacy of this type of heading-normalization is shown in \supsecref{sec:normalizaton}. The second agent-centric network then operates on a per-agent basis. It contains 4 convolutional layers with kernel size 3 and 8 or 16 depth channels. It produces $K \times T \times 5$ parameters describing bivariate Gaussian's per time step per anchor (parameterized by $\mu_x, \mu_y, \log\sigma_x , \log\sigma_y \text{ and } \rho$; the last 3 parameters define the $2\times2$ covariance matrix $\Sigma_{xy}$ in the agent-centric $x,y$-coordinate space), as well as $K$ softmax logits to represent $\pi(\bma|\bmx)$.
\out{
NOTE(bensapp): these are implementation details specific to a particular dataset and should be in the experimental implementation details section, or supplement:

The first scene-level feature extractor is a thinned-out ResNet50 \cite{He16}, with the number of channels reduced to 12.5\% of the default values (also known as using a depth multiplier of 12.5\%). We then extract agent-specific feature patches of size 11px$\times$11px from the scene-level feature map. The patch feature is centered at the agent’s position and rotated according to the agent’s heading. We apply a second CNN on the cropped patch features. This agent-level feature extractor consists of 4 convolutional layers with kernel size 3 each. The depth and stride are, 16/2, 8/2, 16/1, 16/1, respectively.
}
\section{Experiments} \label{sec:experiments}

This section presents empirical results on a number of prediction tasks. We consider the following methods in order to contrast to different aspects of \multiflow.

\PAR{\multiflow $\mu$ [, $\Sigma$].} Our proposed method with multiple anchors, modeling offsets $\mu$ and control uncertainty covariances $\Sigma$. For some experiments, we keep $\Sigma$ frozen, which reduces the maximum-likelihood loss to simple $\ltwo$-loss. However, we can no longer estimate likelihood $p(\bms|\bmx)$ without $\Sigma$ and only report distance-based metrics. 

\PAR{Regression $\mu$ [, $\Sigma$].} To verify our hypothesis that modeling multiple intents is important, we modified the \multiflow architecture to regress a single output trajectory. This is similar to~\cite{Luo18}'s output (but extended to include uncertainty).

\PAR{Min-of-K \cite{Cui19}.} This method predicts $K$ trajectories directly, without pre-defined anchors. The authors define an $\ltwo$-loss on the single trajectory (out of $K$) with minimum distance to the groundtruth trajectory. This is similar to our method, but with implicit anchors and evolving hard-assignment of anchors to groundtruth as training progresses. This representation has inherent ambiguity problems and can suffer from mode collapse. In our experiments below, we extend this method to also predict $\mu, \Sigma$ values at each waypoint to evaluate likelihood.

\PAR{CVAE.} The Conditional Variational Auto-Encoder is a standard implicit generative sampling model and has been successfully adapted to predict trajectory for autonomous driving in \cite{Lee17}. We are interested in comparing its ability to generate a diverse set of samples compared to \multiflow's MAP trajectory per anchor---we hypothesize that \multiflow will have better coverage with the same number of trajectories due to its choice of anchors. For this baseline, we add a CVAE at the end of the second stage agent-centric feature extractor. The decoder and encoder have the same architecture: 4 fully-connected layers of 32 units each.

\PAR{Linear.} Following~\cite{Becker18}, we use a linear model on past states to establish how well a simple constant velocity model can perform.  We fit past observed positions as a linear function of time: $\bmx^t = [\alpha t + \beta, \gamma t + \delta]$ for $t \leq 0$, and use these models to evaluate future positions $\bmx^1, \ldots, \bmx^T$.  We investigated using higher-order polynomials with worse results.

We implemented the single-trajectory regression, Min-of-K, and CVAE using the same input representation and a comparable model architecture in order to achieve a fair comparison. For benchmark datasets, we also report numbers taken from recent publications.

\subsection{Metrics}
\label{sec:metrics}

Different approaches use a variety of output representations; primary examples are single trajectory prediction~\cite{Luo18}, an unweighted set of trajectory samples~\cite{Lee17}, a distribution over trajectories (ours), or probabilistic occupancy grids~\cite{Bansal19}. Each representation comes with its own salient metrics, making it difficult to compare across all methods.  Let $\hat{\mathbf{s}} = \hat{s}_{t = 1 \ldots T}$ be a groundtruth trajectory. We consider the following metrics:

\PAR{Log-likelihood (LL).} We report $\log p(\hat{\bms}|\bmx)$ if the model admits evaluation of likelihood, as does \multiflow when all parameters are learned (see \myeqref{eq:multiflow-likelihood}). The metric is scaled down by a factor of $2 \times T$, where $T$ is the number of time steps and $2$ for the two spatial dimensions.

\out{The log-likelihood is a concise \TODO{chaiy, how to phrase this better} metric compared to the distance-based metrics which often over-penalizes reasonable trajectories that could have happened but did not in the dataset. However, the log-likelihood is less }

\out{\TODO{Do it if time allows.} For methods that generate one or a few discrete trajectories without estimating covariances, we find a single scalar $\alpha$ to represent variances at all waypoints. We assume independence between and within waypoints. For generative methods where a large number of independent samples can be drawn, we apply kernel-density-estimation (KDE) on the samples to yield a mixture of Gaussian, which in turn estimates the log-likelihood of a groundtruth trajectory. The log-likelihood is our {\em preferred} metric for the ablation. Unfortunately, the log-likelihood approximated via post-processing may suffer from great approximation error and become less comparable. The log-likelihood has the unit in nat.
}

\PAR{Distance-based.} In this category are the commonly-used average displacement error ({\bf ADE}) $\frac{1}{T} \sum_{t=1}^T \left \| \hat{s}_t - s^*_t \right \| _2$ and final displacement error ({\bf FDE}) $\left \| \hat{s}_T - s^*_T \right \| _2$,  where $\mathbf{s}^*$ is the most-likely within a weighted set. For evaluating a set of trajectories, {\bf minADE$_M$} $\min_{s_m} \frac{1}{T} \sum_{t=1}^T \left \| \hat{s}_t - s_{m, t} \right \| _2$ measures the displacement error against the closest trajectory in the set of size $M$, so that reasonable predictions that simply do not happen to be the logged groundtruth are not penalized. Note that there is also the {\bf minMSD$_M$} \cite{Rhinehart18}, which is similar but the average is calculated on squared distances instead.

% {\bf meanADE$_M$} takes the average over $M$ predictions instead of taking the closest.
% If we consider probabilities associated with predictions, we can generalize meanADE to a weighted-ADE ({\bf wADE}) $\sum_{m} w_k \frac{1}{T} \sum_{t=1}^T \left \| \hat{s}_t - s_{m, t} \right \| _2 / \sum_m w_m$. 

\subsection{Toy experiment: 3-way intersection}
\label{sec:toy}

\begin{figure}[!tbp]
\centering
\includegraphics[width=0.99\textwidth]{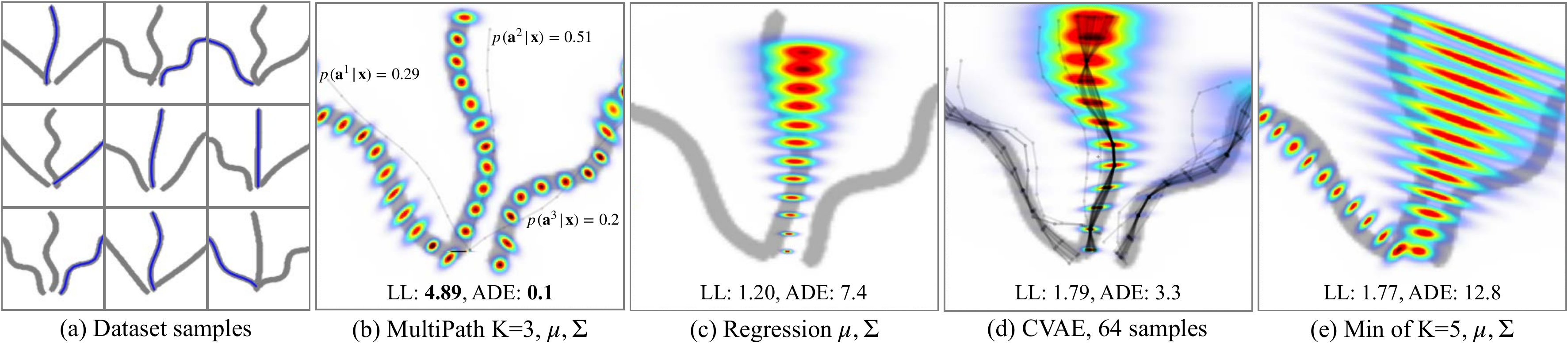}
\caption{\small Results on the 3-way intersection toy example. Due to large dynamic range, uncertainties shown with jet colormap are scaled per-plot per-timestep, and not directly comparable. (a) Samples drawn from the data generation procedure, with groundtruth paths shown as blue lines. (b) \multiflow with $K=3$ anchors correctly learns the intent and uncertainty distributions, achieving high likelihood. The anchors (light gray) were estimated by averaging $10^5$ samples. (c) Single-trajectory modeling predicts a mean over all paths with corresponding location uncertainty which grows as paths diverge. (d) CVAE samples with a Kernel Density Estimate distribution fit~\cite{Bishop06}. (e) Min-of $K=5$ trajectories.  This model is very sensitive to initial weights, and on 5 trials with 4 learning rates, collapsed to only 1 or 2 active modes only (2 are shown here).  We initialized starting regression weights to be uniform random in a small region surrounding the $t=0$ position, for a better chance of learning multiple unique modes.
\vspace{-0.2cm}}
\label{fig:toy-3way}
\end{figure}

We first explore a simple proof-of-concept dataset generated based on our modeling assumptions.  We generate synthetic 3-way intersections, with the probability of choosing the left, the middle or the right path set a priori to be the {\em intent uncertainty} distribution  $\{0.3, 0.5, 0.2\}$. To emphasize the flexibility of our single-trajectory {\em control uncertainty} modeling, each path is generated by sampling parameterized sine waves: $y = \sin(\omega t + \phi)$, where the frequency $\omega \sim \mathcal{U}(0,2)$ and phase shift $\phi \sim \mathcal{U}(-\pi, \pi)$.  As shown in Figure~\ref{fig:toy-3way}, \multiflow is able to fit the underlying distribution correctly, recovering the intent uncertainty, and reaching approximately Bayes-optimal likelihood, while other methods fare worse.

\subsection{Behavior prediction for autonomous driving.}
\label{sec:bp}

\begin{table}[!tbp]
\caption{\small Comparison of \multiflow on a large autonomous driving dataset. Brackets appear for cells where a metric expects a set of future trajectories but the method only produces one. All experiments (except for {\em linear}) were repeated 5 times with random initialization to produce the mean and standard deviation values. \textbf{LL} is the log-likelihood. \textbf{ADE} is the average distance error. {\bf minADE$_M$} is the top-$M$ average distance error given $M$ trajectory predictions. One out of five CVAE runs degenerated. \textbf{CVAE} includes all 5 runs, while \textbf{CVAE select} excludes the degenerated run. A more detailed analysis is in \supsecref{sec:analysis}.}
\label{tbl:driving-sota}
\centering
\begin{tabular}{lccccc}
\toprule
\multirow{1}{*}{Method} & LL $\uparrow$ & ADE $\downarrow$ & minADE$_5$ $\downarrow$ & minADE$_{10}$ $\downarrow$ & minADE$_{15}$ $\downarrow$ \tabularnewline
\midrule
Linear & - & 3.26 & (3.26) & (3.26) & (3.26) \tabularnewline
Regression $\mu$ & - & \textbf{1.17\tiny{$\pm$0.01}} & (1.17{\tiny$\pm$0.01} ) & (1.17{\tiny$\pm$0.01} ) & (1.17{\tiny$\pm$0.01} ) \tabularnewline
Regression $\mu,\Sigma$ & 3.64\tiny{$\pm$0.01} & 1.41\tiny{$\pm$ 0.02} & (1.41{\tiny $\pm$ 0.02}) & (1.41\tiny{$\pm$ 0.02} ) & (1.41\tiny{$\pm$ 0.02} ) \tabularnewline
CVAE & - & 2.16\tiny{$\pm$2.15} & 1.82\tiny{$\pm$2.35}  & 1.74\tiny{$\pm$2.39}  & 1.71\tiny{$\pm$2.41}  \tabularnewline
CVAE select & - & 1.20\tiny{$\pm$0.03} & 0.77\tiny{$\pm$0.03} &  0.67\tiny{$\pm$0.03} &  0.63\tiny{$\pm$0.03} \tabularnewline
Min-Of-K $\mu$ \cite{Cui19} & - & 1.37\tiny{$\pm$0.02} & 0.87{\tiny$\pm$0.02} & 0.86{\tiny$\pm$0.02} & 0.86{\tiny$\pm$0.02} \tabularnewline
Min-Of-K $\mu, \Sigma$ \cite{Cui19} & 4.26\tiny{$\pm$0.04} & 1.23\tiny{$\pm$ 0.02} & 0.70{\tiny $\pm$ 0.01} & 0.70\tiny{$\pm$ 0.01} & 0.70\tiny{$\pm$ 0.01} \tabularnewline
MultiPath $\mu$ (ours) & - & \textbf{1.17\tiny{$\pm$0.00}} & \textbf{0.58\tiny{$\pm$0.00}} & \textbf{0.48\tiny{$\pm$0.00}} & \textbf{0.46\tiny{$\pm$0.00}} \tabularnewline
MultiPath $\mu,\Sigma$ (ours) & \textbf{4.37\tiny{$\pm$0.00}} & 1.25\tiny{$\pm$0.01} & 0.63\tiny{$\pm$0.00} &  0.61\tiny{$\pm$0.00} &  0.61\tiny{$\pm$0.00} \tabularnewline
\bottomrule
\vspace{-0.2cm}
\end{tabular}
\end{table}

To verify the performance of the proposed system, we collected a large dataset of real-world driving scenes from several cities in North America. Data is captured by a vehicle equipped with cameras, lidar and radar. As in~\cite{Casas18,Hong19}, we assume that an industry-grade perception system provides sufficiently accurate poses and tracks for all nearby agents, including vehicles, pedestrians, and cyclists. In our experiments, we treat the sensing vehicle as an additional agent, indistinguishable from any other agent in the scene.
Most of the collected vehicle trajectories are either stationary or moving straight at a constant speed. Neither case is particularly interesting from a behavior prediction point of view.  To address this and other dataset skew, we partitioned the space of future trajectories via a uniform, 2D grid over constant curvatures and distances, and performed stratified sampling such that the number of examples in each partition was capped to be at most 5\% of the resulting dataset. The balanced dataset totals 3.85 million examples, contains 5.75 million agent trajectories and constitutes approximately 200 hours of driving.

The top-down rendered input tensor for this data has a resolution of 400 px $\times$ 400 px and corresponds to 80 m $\times$ 80 m in real-world coordinates. We sample time steps every 0.2 s (5 Hz). The following features are stacked in the depth dimension: 3 channels of color-coded road semantics, 1 channel of distance-to-road-edge map, 1 channel encoding the speed limit, 5 channels encoding the traffic light states over the past 5 time steps (=1 second), and 5 channels each showing vehicles' top-down orthographic projection for each of the past 5 time steps. This results in 15 input channels in total. We predict trajectories up to 30 frames / 6 seconds into the future. The number of anchors $K$ is set to 16 for \multiflow $\mu, \Sigma$ and 64 for \multiflow $\mu$. The scene-level network is a ResNet50 \cite{He16} with a depth multiplier of 25\%, followed by a depth-to-space operation that restores some of the lost spatial resolution in the ResNet back to $200\times200$. Finally, we train the model end-to-end for 500k steps at a batch size of 32, with a learning rate warm-up phase and a cosine learning rate decay 

Experimental results are shown in \tblref{tbl:driving-sota}. \multiflow outperforms the baselines in all metrics. With respect to the log-likelihood, we have observed the most log-likelihood measurements for this task to fall between 3 to 4.2 nats, so the gain of roughly 0.2 nat by \multiflow compared to the regression baseline is quite significant. See \supsecref{sec:analysis} for in-depth analyses of these results. 16 anchors are used for \multiflow $\mu, \Sigma$, while 64 was the best $K$ for \multiflow $\mu$. An analysis of the effect of the number of anchors $K$ is in \supsecref{sec:anchors}, while the figures in \supsecref{sec:viz_clusters} visualize the anchors.

\begin{figure}[tp]
  \centering
\includegraphics[width=\textwidth]{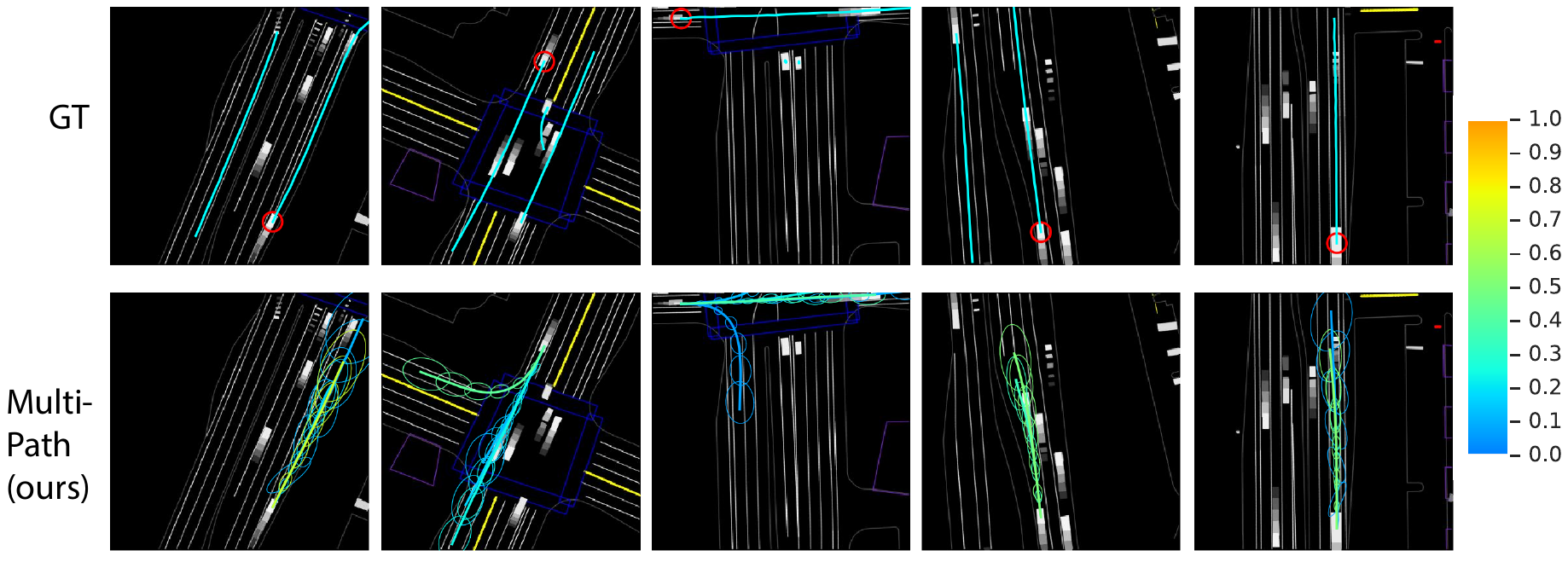}
\caption{\small \textbf{\multiflow example results.}
We show in one column the result for one agent (focused agent) in a scene.
\textbf{Top}: Logged trajectories of all agents are displayed in cyan. The focused agent is highlighted by a red circle. \textbf{Bottom}: \multiflow showing up to 5 trajectories with uncertainty ellipses. Trajectory probabilities (softmax outputs) are encoded in a color map shown to the right. \multiflow can predict uncertain future trajectories for various speed (1st column), different intent at intersections (2nd and 3rd columns) and lane changes (4th and 5th columns), where the regression baseline only predicts a single intent.
\vspace{-0.4cm}
}
\label{fig:examples}
\end{figure}

\subsection{Stanford Drone}
\label{sec:sdd}

\begin{table}[!tbp]
\caption{\small Comparison of \multiflow and baselines on the Stanford Drone Dataset (SDD). Distance measures are in terms of pixels in the original video resolution.}
\label{tbl:sdd}
\centering
\begin{tabular}{lcccc}
\toprule
Method & LL $\uparrow$ & ADE $\downarrow$  & FDE $\downarrow$ & minADE$_5$ $\downarrow$ \\
\midrule          
Linear                               & --    & 26.14 & 53.24 & --           \\
CVAE                                 & --    & 30.91 & 61.40 & 26.29        \\
Regression $\mu$                     & --    & 27.44 & 56.44 & --           \\
Regression $\mu, \Sigma$             & 3.06  & 26.67 & 54.34 & --           \\
\multiflow $\mu, \Sigma$             & \textbf{3.52} & 28.32 & 58.38 & \textbf{17.51}        \\
\midrule                                 
DESIRE-SI-IT0~\cite{Lee17}           & --    & 36.48 & 61.35 & 30.78        \\
CAR-Net~\cite{Sadeghian18}           & --    & \textbf{25.72} & \textbf{51.80} & --           \\
Social Forces~\cite{Yamaguchi2011}   & --    & 36.48 & 58.14 & --           \\
Social LSTM~\cite{Alahi16}           & --    & 31.19 & 56.97 & --           \\
\bottomrule
\vspace{-0.2cm}
\end{tabular}
\end{table}

The Stanford Drone Dataset~\cite{Robicquet16} consists of top-down, near-orthographic videos of college campus scenes, collected by drones, containing interacting pedestrians, cyclists and vehicles.  The RGB camera frames provide context similar to a rendered road semantics in the driving vehicle environment, and we treat it as such.  We use the most common settings in the literature: sampling at 2.5 Hz, and predicting 4.8 seconds (12 frames) into the future, using 2 seconds of history (5 frames).  Additional experimental details are in \supsecref{sec:sdd_experiment_setup}.

As shown in ~\tblref{tbl:sdd}, we perform at or better than state-of-the-art in best single-trajectory distance metrics.  Notably, CAR-Net~\cite{Sadeghian18} outperforms our comparable single-trajectory model; their method focuses on a sophisticated attention and sequential architecture tuned to get the best single-trajectory distance metric performance.  Interestingly, our single-trajectory model performs better when trained to predict uncertainty as well, a potential benefit of modeling uncertainty discussed in~\cite{Kendall17}.  

\subsection{CARLA}
\label{sec:carla}

\newcolumntype{H}{>{\setbox0=\hbox\bgroup}c<{\egroup}@{}}

% mflow_carla_strgt_anc_bev12_res50_100x100_unc_lr_0.01
\begin{table}[!tbp]
\caption{\small Comparison of \multiflow and baselines on the CARLA Dataset. {\bf minMSD$_{12}$} is the minimum mean-squared-distance computed on the top 12 predicted trajectories.}
\label{tbl:carla}
\centering
\begin{tabular}{lHc}
\toprule
\textbf{Town01 (5 agents)} & LL $\uparrow$ & minMSD$_{12}$ $\downarrow$  \\
\midrule          
DESIRE~\cite{Lee17}                 & --       & 2.60 \\
SocialGAN~\cite{Gupta19}            & --       & 1.46 \\
R2P2-MA~\cite{Rhinehart18}          & 0.630    & 0.84 \\
ESP~\cite{Rhinehart19}              & 0.643    & 0.72 \\
\multiflow $\mu, \Sigma$            & --       & {\bf 0.68} \\
\bottomrule
\vspace{-0.2cm}
\end{tabular}
\quad
\begin{tabular}{lHc}
\toprule
\textbf{Town02 (5 agents)} & LL $\uparrow$ & minMSD$_{12}$ $\downarrow$  \\
\midrule
DESIRE~\cite{Lee17}                 & --        & 2.42 \\
SocialGAN~\cite{Gupta19}            & --        & 1.14 \\
R2P2-MA~\cite{Rhinehart18}          & 0.618     & 0.77 \\
ESP~\cite{Rhinehart19}              & 0.630     & {\bf 0.68} \\
\multiflow $\mu, \Sigma$            & --        & 0.69 \\
\bottomrule
\vspace{-0.2cm}
\end{tabular}
\end{table}

We evaluate \multiflow on the publicly available multi-agent trajectory forecasting and planning dataset generated using the CARLA~\cite{Dosovitskiy17} simulator by \cite{Rhinehart19}. Experimental details are in \supsecref{sec:carla_experiment_setup}.
\tblref{tbl:carla} reproduces results reported by \cite{Rhinehart19} for the DESIRE~\cite{Lee17}, SocialGAN~\cite{Gupta19}, R2P2-MA~\cite{Rhinehart18}, and the PRECOG-ESP~\cite{Rhinehart19} methods and compares the performance of \multiflow against them. We report the minMSD metric with the top $K=12$ predictions as defined in \cite{Rhinehart19} to report our evaluation results.

%We report the Log-Likelihood and minMSD (with $K=12$) metrics as defined in \cite{Rhinehart19} to report our evaluation results. \TODO{State why and what our results convey.}

\section{Conclusion}
We have introduced \multiflow, a model which predicts parametric distributions of future trajectories for agents in real-world settings.  Through synthetic and real-world datasets, we have shown the benefits of \multiflow over previous single-trajectory and stochastic models in achieving likelihood and trajectory-set metrics and needing only 1 feed-forward inference pass.

\small{\acknowledgments{We like to thank Anca Dragan, Stephane Ross and Wei Chai for their helpful comments.}}

\small {
\bibliography{ms}  % .bib
}

\newpage

\appendix

\section{In-depth analysis}
\label{sec:analysis}

We provide a more detailed analysis of \tblref{tbl:driving-sota}. The results are strictly different representations of the same experiments, all evaluated on the autonomous driving dataset, as described in \secref{sec:bp}.

\subsection{Group by groundtruth final waypoint}

Statistically, most vehicles we observe are parked (stationary) or going straight. Neither case is particularly interesting since a trivial solution such as the linear fitting is probably good enough to solve them. In this exercise, we split the evaluation samples into seven disjoint categories according to the location of the final waypoint at 6 seconds: 1) \textbf{Stationary}, agents that have moved less than 4 meters; 2) \textbf{Slow}, agents that moved farther than 4 meters but less than 8 meters; 3) \textbf{Straight}, agents that moved farther than 8 meters and the final waypoint is within $\pm 5^{\circ}$ with respect of the initial heading; 4) Slight left (\textbf{SLeft}), farther than 8 meters and between $-5^{\circ}$ and $-30^{\circ}$; 5) \textbf{Left}, farther than 8 meters and larger than $-30^{\circ}$; 6) Slight right (\textbf{SRight}), same as SLeft, but in the other direction; 7) \textbf{Right}, same as Left, but in the other direction.

\begin{figure*}[!htbp]
\centering
  \begin{subfigure}[b]{0.49\textwidth}
    \includegraphics[width=\textwidth]{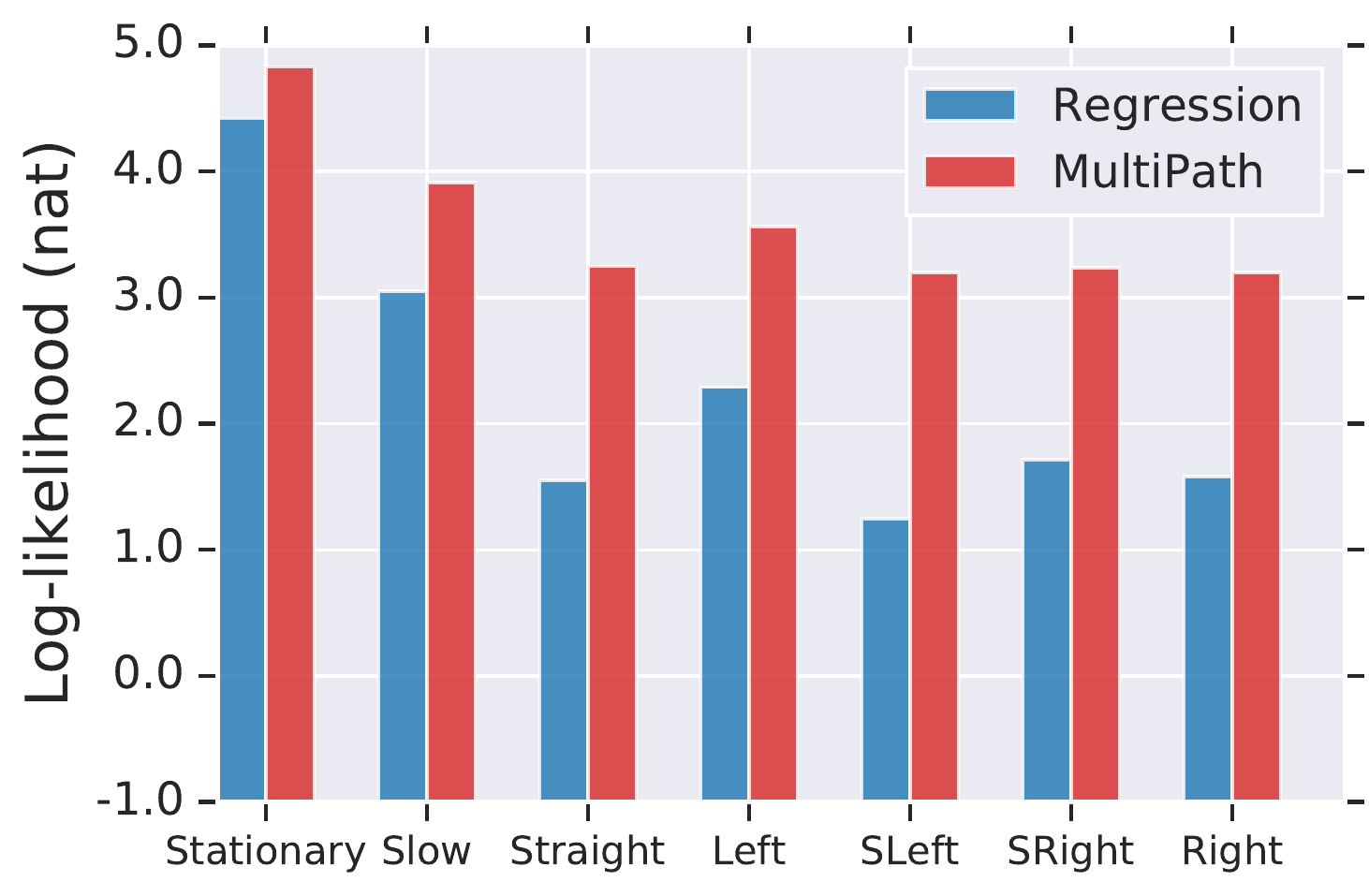}
    \caption{Log-likelihood}
    \label{fig:finer_ll}
  \end{subfigure}
  \begin{subfigure}[b]{0.49\textwidth}
    \includegraphics[width=\textwidth]{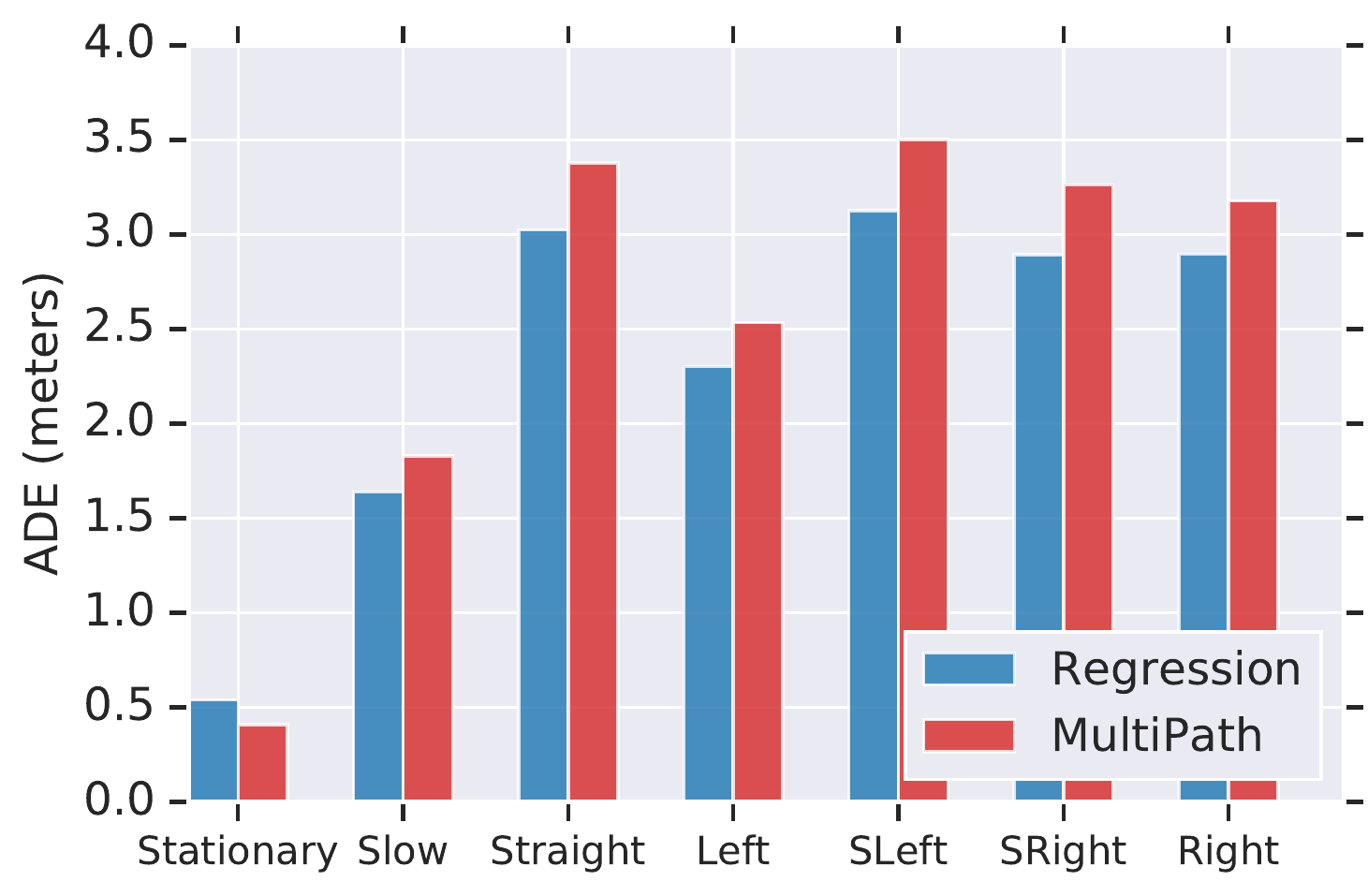}
    \caption{ADE}
    \label{fig:finer_ade}
  \end{subfigure}
  \begin{subfigure}[b]{0.49\textwidth}
    \includegraphics[width=\textwidth]{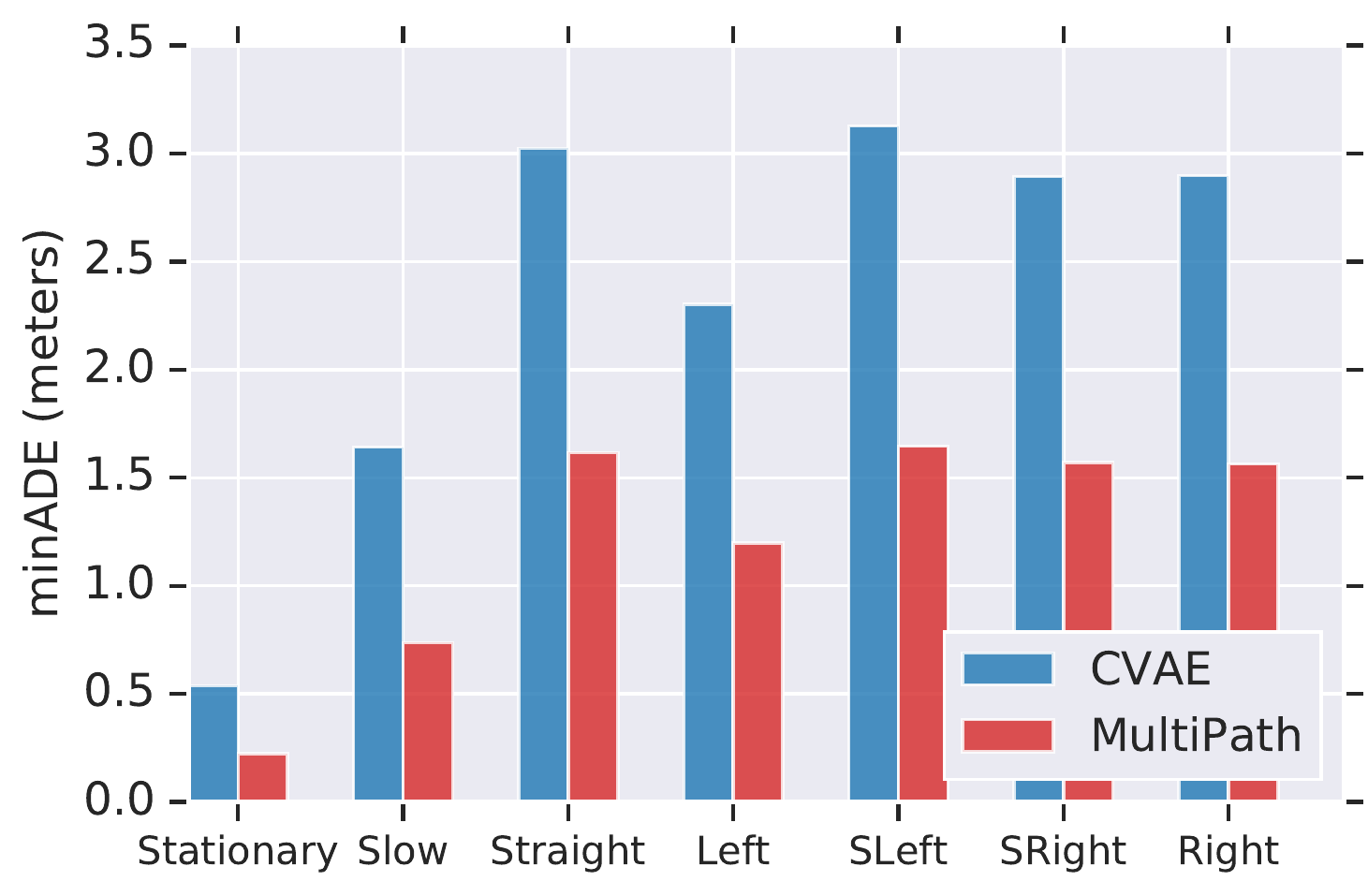}
    \caption{meanADE}
    \label{fig:finer_cvae}
  \end{subfigure}  
\caption{
Detailed results grouped by the groundtruth final waypoint. (a): Detailed comparison between \multiflow $\mu,\Sigma$ and the regression $\mu,\Sigma$ baseline on the log-likelihood. (b): Detailed comparison between \multiflow $\mu$ and the regression $\mu$ baseline on the ADE. Detailed comparison between \multiflow $\mu$ and the CVAE baseline on the meanADE. The top 5 trajectories are kept for \multiflow, and 100000 trajectories are sampled from the CVAE.
}
\label{fig:finer}
\end{figure*}

Results are shown in \figref{fig:finer}. In \figref{fig:finer_ll}, we compare \multiflow $\mu,\Sigma$ against the regression baseline $\mu,\Sigma$ on the log-likelihood metric. As expected, the baseline performs reasonably well in the Stationary and Slow categories but is unable to model faster trajectories, where \multiflow outperforms the baseline decisively.

\subsection{Group by time step}

\begin{figure*}[!htbp]
\centering
\includegraphics[width=0.35\textwidth]{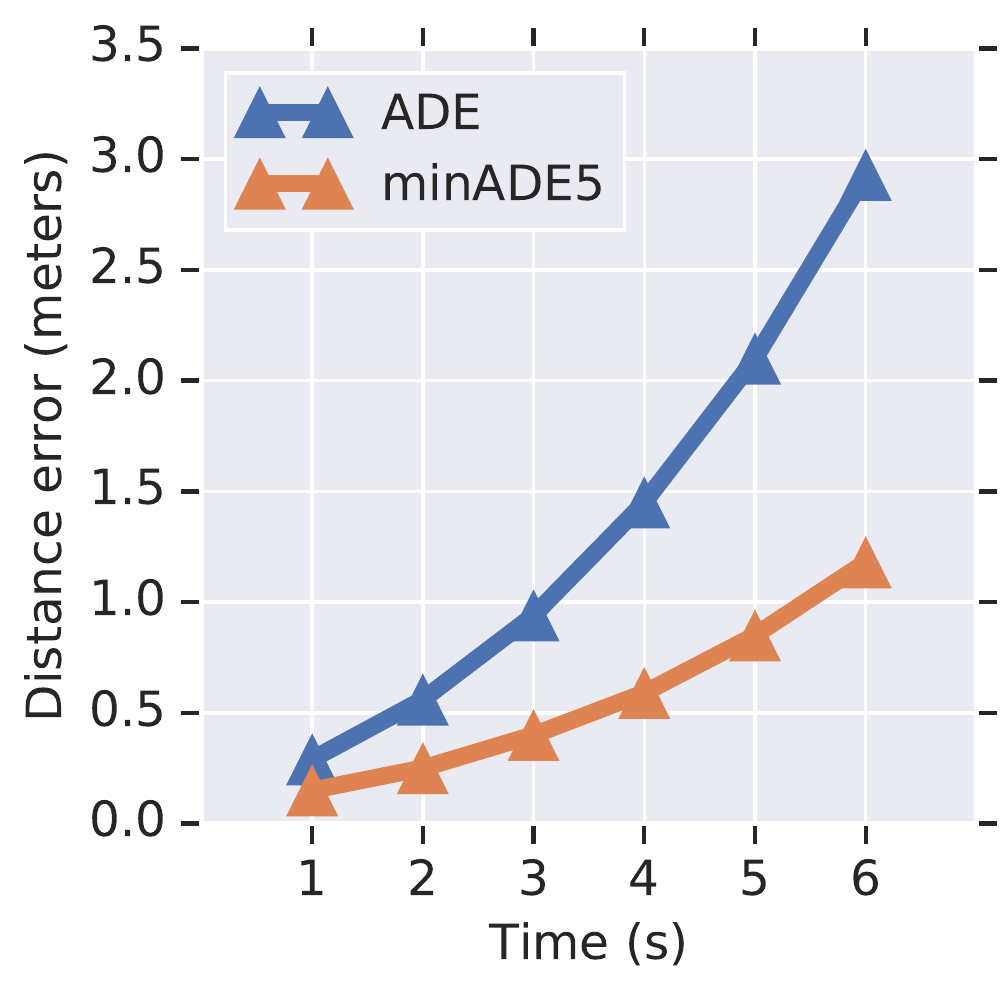}
\caption{
Detailed results from \multiflow $\mu$ grouped by time step on the ADE and minADE$_{5}$ metrics.
}
\label{fig:steps_l2}
\end{figure*}

We look at the distance-based errors as the function of the predicted time step up to 6 seconds in \figref{fig:steps_l2}. This error is usually expected to grow linearly in time. However, our curves are slightly super-linear, most likely caused by the natural distribution of heading-normalized trajectories, where the acceleration in both heading and speed is smooth.

\section{Hyperparameters}
\label{sec:hyperparameters}

There are numerous moving parts of the proposed model. In this section, we conducted extensive experiments on the choice of key hyperparameters, discuss the sensitivity of the model with respect to these parameters.

\subsection{Number of anchors $K$}
\label{sec:anchors}

\begin{figure*}[!htbp]
\centering
  \begin{subfigure}[b]{0.35\textwidth}
    \includegraphics[width=\textwidth]{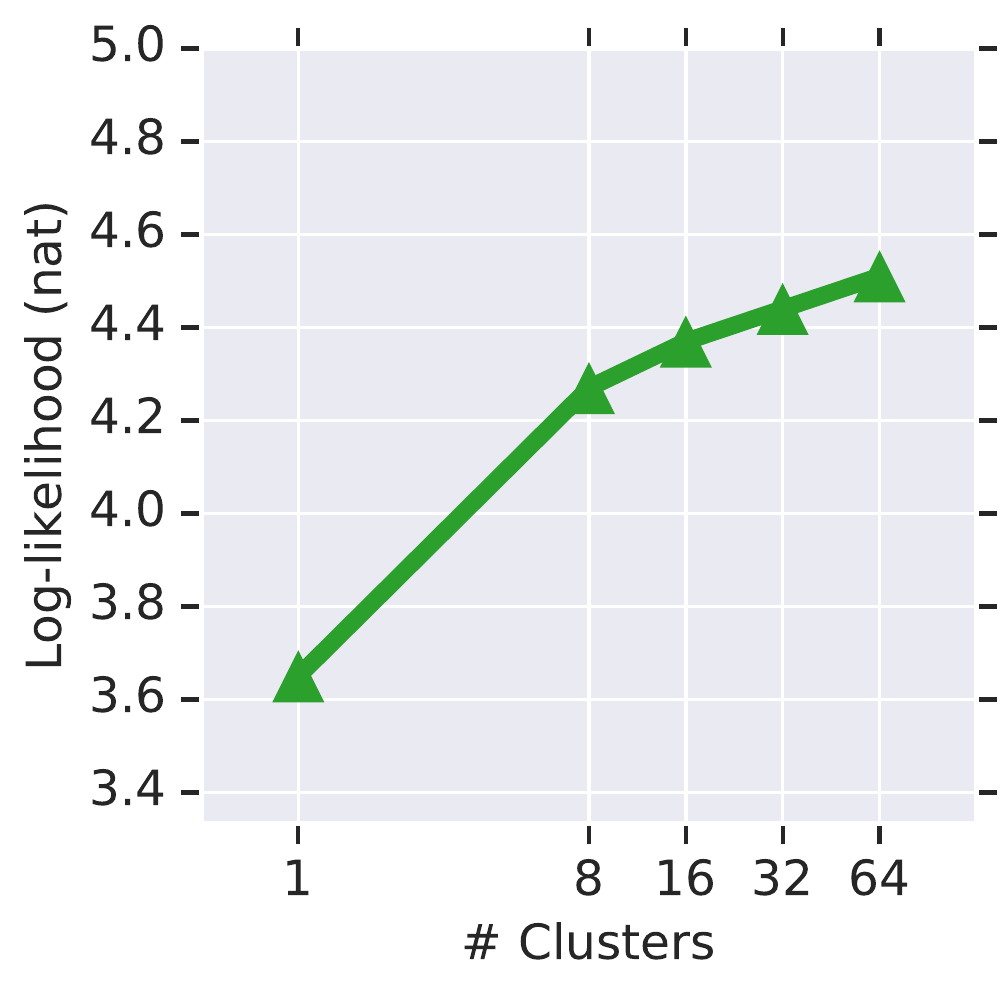}
    \caption{Log-likelihood}
    \label{fig:clusters_ll}
  \end{subfigure}
  \begin{subfigure}[b]{0.35\textwidth}
    \includegraphics[width=\textwidth]{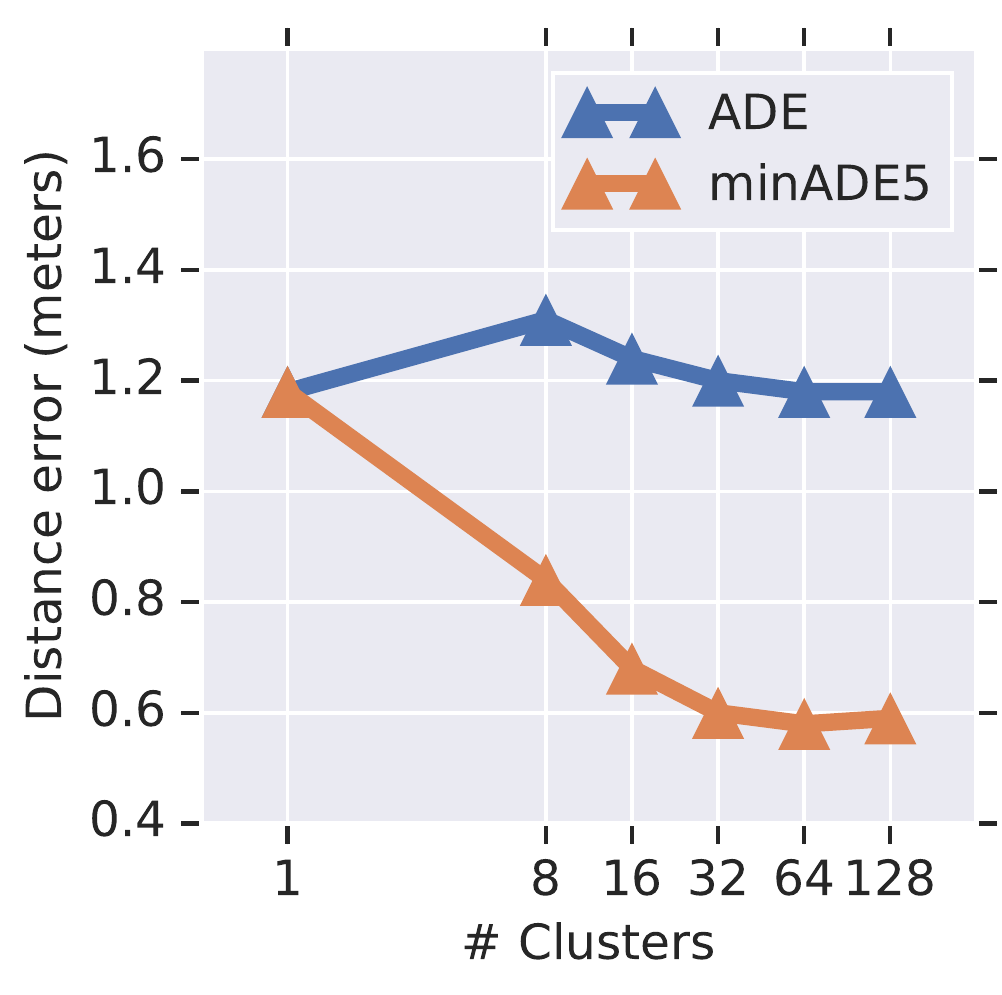}
    \caption{ADE/minADE$_5$}
    \label{fig:clusters_l2}
  \end{subfigure}
\caption{Results as function of the number of clusters $K$. (a): \multiflow $\mu,\Sigma$ on the log-likelihood metric with $K$ ranging from 1 to 64. (b): \multiflow $\mu$ on the ADE and minADE$_5$ metrics, with $K$ ranging from 1 to 128.
}
\label{fig:clusters}
\end{figure*}

One of the key hyperparameters of \multiflow is the choice of the number of anchors $K$, as it is always the case with methods that rely on unsupervised clustering. To this end, we trained several \multiflow $\mu,\Sigma$ and \multiflow $\mu$ models with different $K$s, the results are shown in \figref{fig:clusters}.

\subsection{Backbone network sensitivity}
\label{sec:backbone}

\begin{figure*}[!htbp]
\centering
  \begin{subfigure}[b]{0.35\textwidth}
    \includegraphics[width=\textwidth]{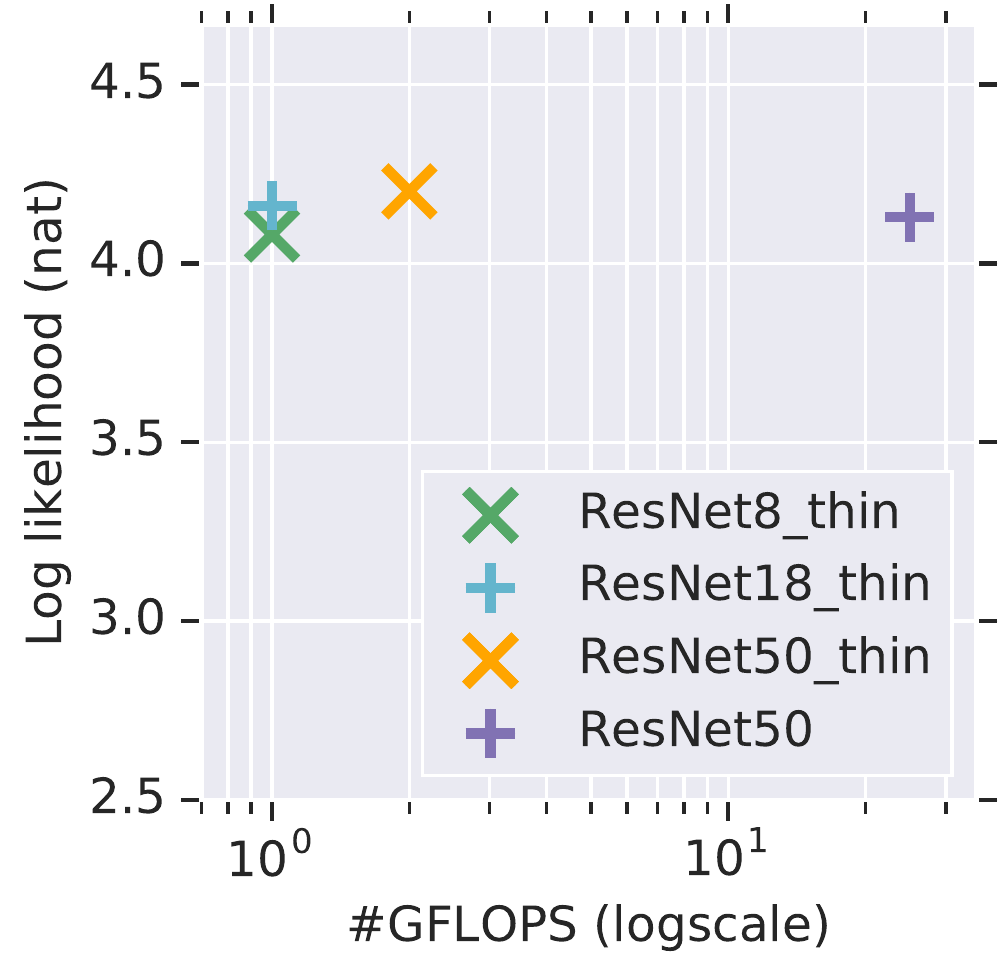}
    \caption{GFLOPS}
    \label{fig:archs_flops}
  \end{subfigure}
  \begin{subfigure}[b]{0.35\textwidth}
    \includegraphics[width=\textwidth]{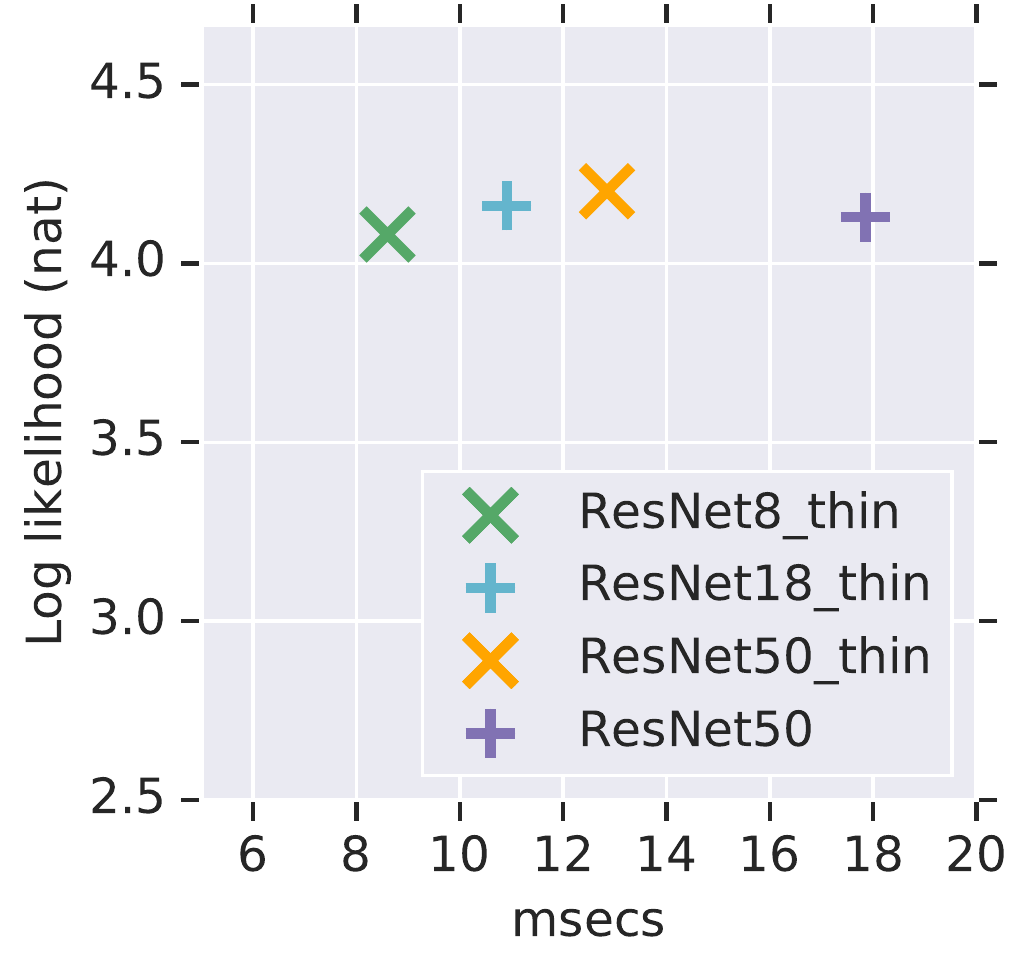}
    \caption{msecs}
    \label{fig:archs_msecs}
  \end{subfigure}
\caption{Analysis of different backbone networks. a) Theoretical complexity in GFLOPS (1 billion FLOPS). b) Empirical latency on a TITAN GPU, averaged over 100 runs. The ResNet50\_thin backbone is used in all other experiments. Complexity and latency are measured assuming exactly 10 agents per scene.
}
\label{fig:archs}
\end{figure*}

We run the \multiflow $\mu, \Sigma$ model using 4 ResNet \cite{He16} backbones of various complexity: ResNet8\_thin, ResNet18\_thin, ResNet50\_thin, ResNet50. The backbones denoted with {em thin} are using a depth multiplier of 25\%. The results are shown in \figref{fig:archs}.

\subsection{Importance of heading normalization}
\label{sec:normalizaton}

\begin{figure*}[!htbp]
\centering
  \begin{subfigure}[b]{0.32\textwidth}
    \includegraphics[width=\textwidth]{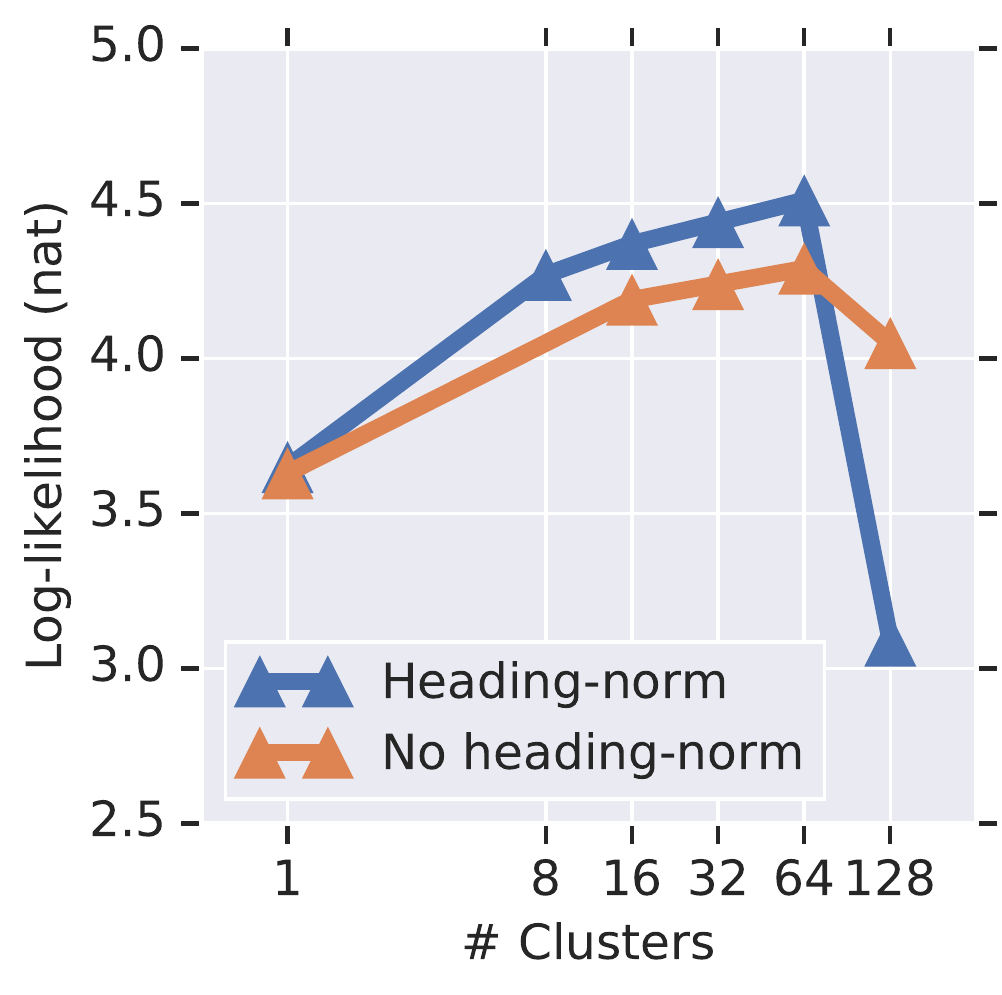}
    \caption{Log-likelihood}
    \label{fig:heading_norm_ll}
  \end{subfigure}
  \begin{subfigure}[b]{0.32\textwidth}
    \includegraphics[width=\textwidth]{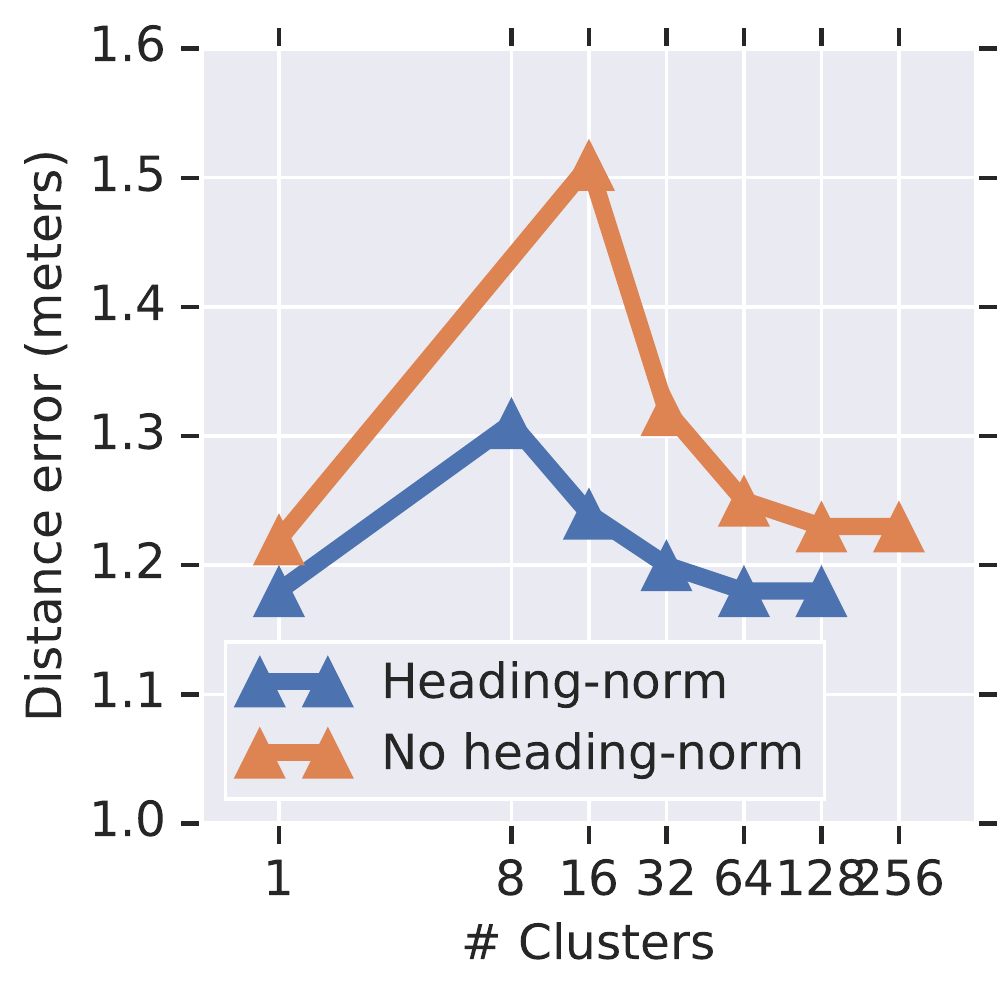}
    \caption{ADE}
    \label{fig:heading_norm_ade}
  \end{subfigure}
  \begin{subfigure}[b]{0.32\textwidth}
    \includegraphics[width=\textwidth]{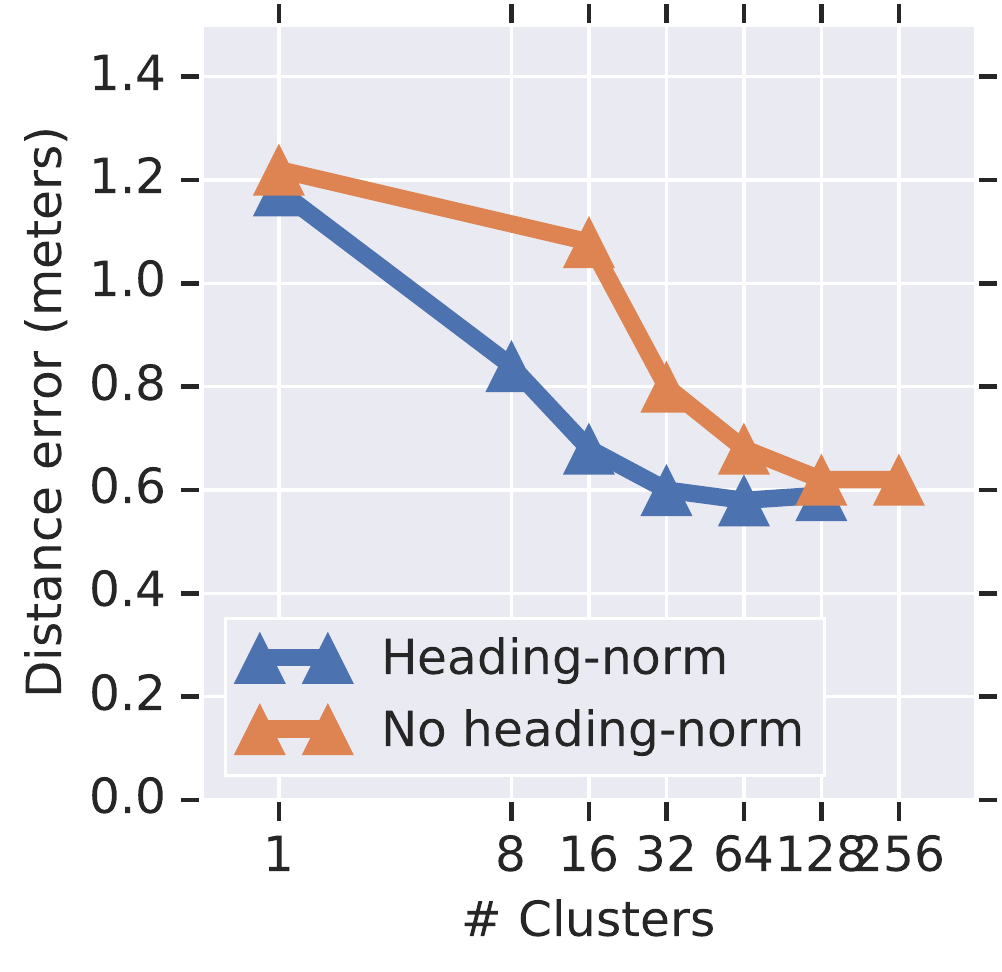}
    \caption{minADE$_5$}
    \label{fig:heading_norm_minade}
  \end{subfigure}  
\caption{Impact of heading normalization on three metrics. Having the heading compensated during agent-centric feature crop is beneficial throughout.
}
\label{fig:heading}
\end{figure*}

The heading of a vehicle is a valuable signal, especially since cars cannot move in arbitrary directions. In \secref{sec:bp}, we assumed that the heading of the agents is given, alongside the positions and sizes of them over the past 1 second. We then crop the agent-specific feature map while compensating for the heading, so that the heading of the agent points to the same direction after the crop. In \figref{fig:heading}, we show the importance of this heading normalization.

As expected, the sweet spot for the number of clusters $K$ larger when the heading is not compensated for. It appears that empirically, the heading is of great importance, and the heading normalized models outperform their counterparts on all three metrics.

\subsection{Road semantic information and traffic lights}
\label{sec:roadgraph}

One main benefit of using a rasterized top-down input representation is the simplicity to encode spatial information such as semantic road information and traffic lights, which are crucial for human driving. To validate that \multiflow also considers this contextual information, we trained ablation models by stripping the road information and traffic lights from the input. Results are shown in \tblref{tbl:context}. The gain of including the contextual information is significant in all distance-based metrics. Interestingly, the gain is less apparent in the log-likelihood. We believe this is due to the fact that the context is mostly useful to prevent false positive trajectories, which has a smaller impact on the log-likelihood.

\begin{table}[!htbp]
\caption{Impact of including the semantic road information and traffic light context.}
\label{tbl:context}
\centering
\begin{tabular}{lccccc}
\toprule
\multirow{1}{*}{Method} & Log-likelihood $\uparrow$ & ADE $\downarrow$ & FDE $\downarrow$ & minADE$_5$ $\downarrow$ \tabularnewline
\midrule
With context $\mu,\Sigma$ & 4.37 & 1.25 & 3.17 & 0.63  \tabularnewline
No context $\mu,\Sigma$ & 4.32 & 1.45 & 3.83 & 0.68 \tabularnewline
\midrule 
With context $\mu$ & - & 1.18 & 2.93 & 0.58 \tabularnewline
No context $\mu$ & - & 1.44 & 	3.71 & 	0.73 \tabularnewline
\bottomrule
\end{tabular}
\end{table}

\section{Visualization of all anchors}
\label{sec:viz_clusters}

\begin{figure*}[!htbp]
\centering
  \begin{subfigure}[b]{0.23\textwidth}
    \includegraphics[width=\textwidth]{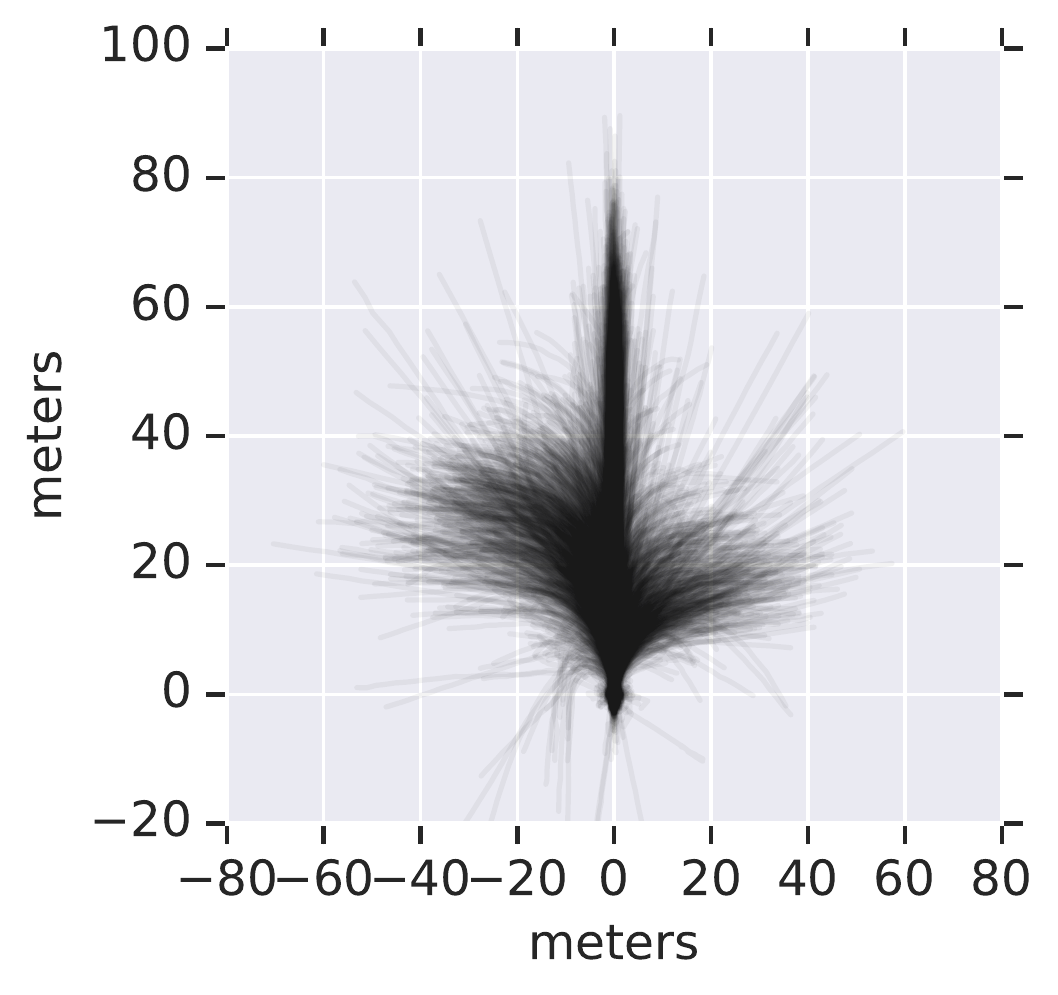}
    \caption{Trajectories}
  \end{subfigure}
  \begin{subfigure}[b]{0.23\textwidth}
    \includegraphics[width=\textwidth]{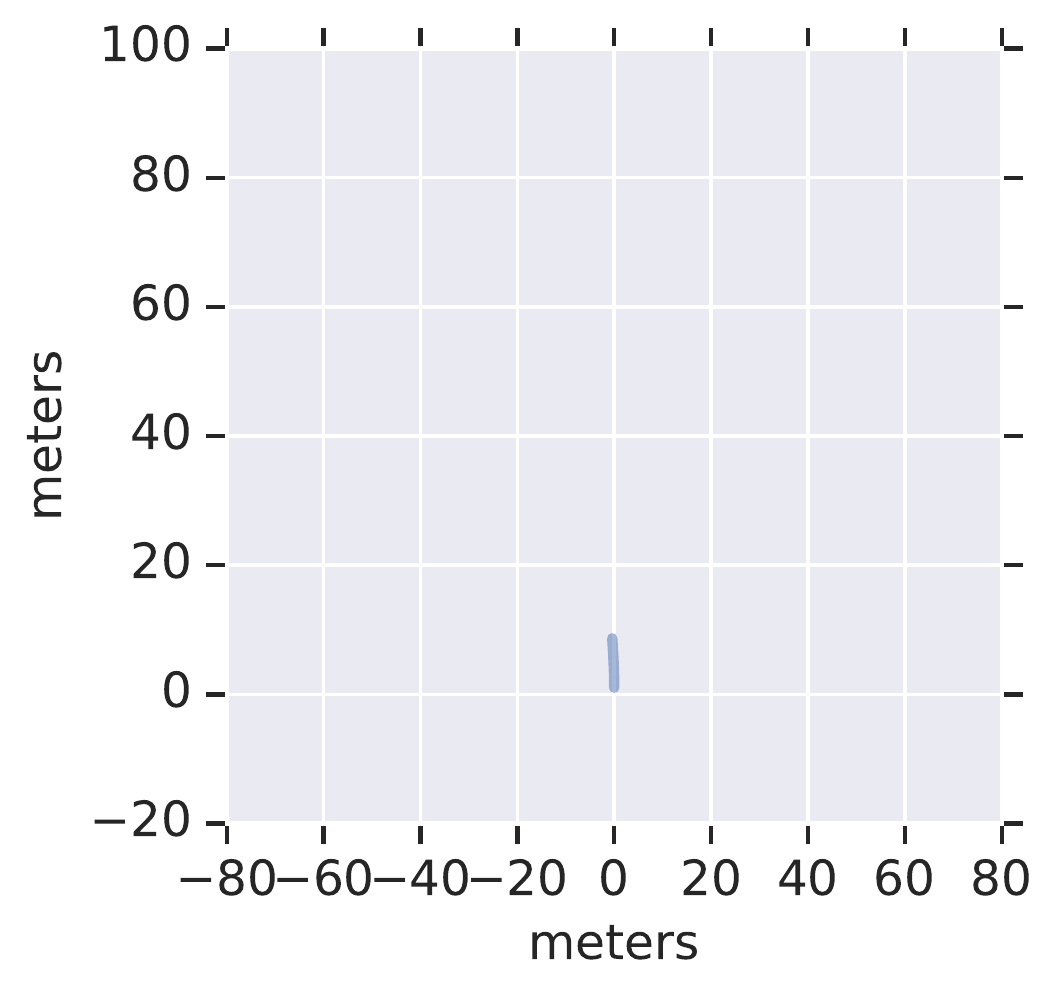}
    \caption{$K=1$}
  \end{subfigure}
  \begin{subfigure}[b]{0.23\textwidth}
    \includegraphics[width=\textwidth]{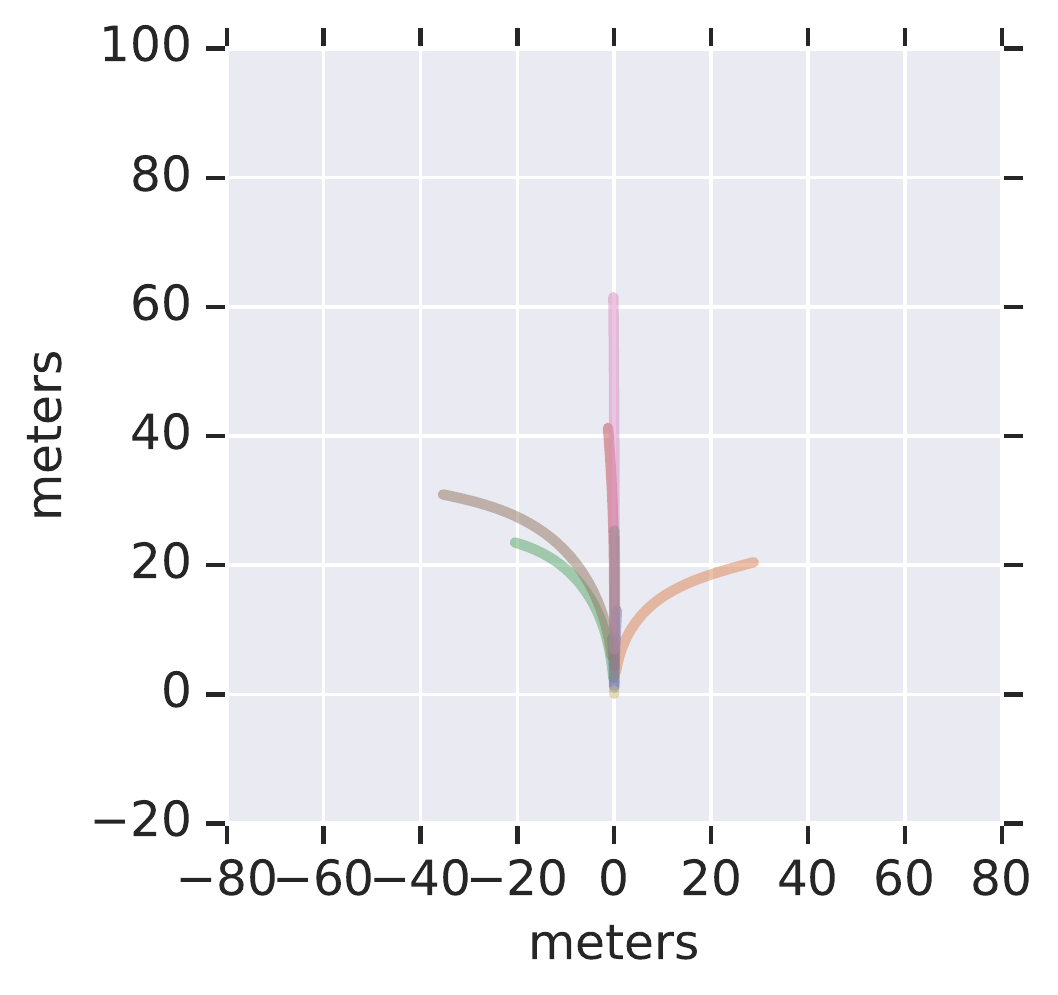}
    \caption{$K=8$}
  \end{subfigure}  
  \begin{subfigure}[b]{0.23\textwidth}
    \includegraphics[width=\textwidth]{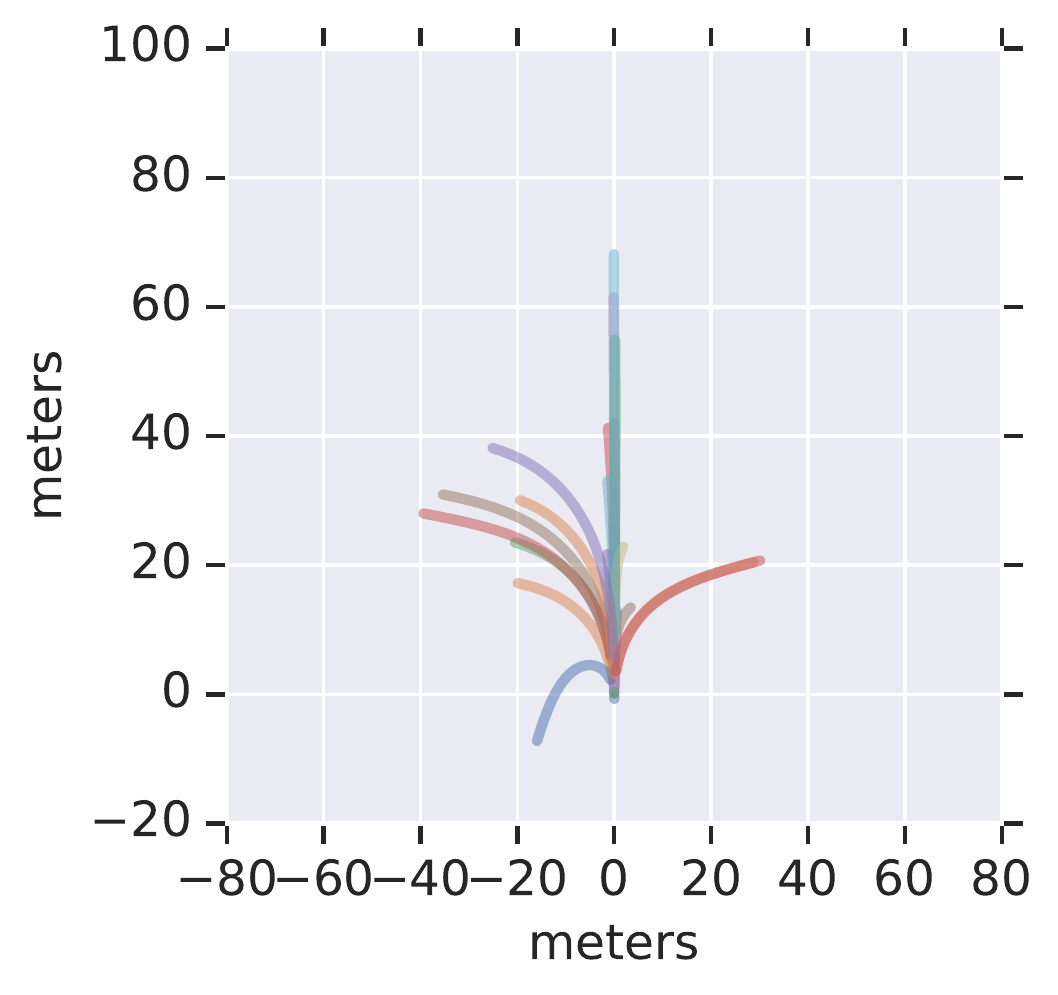}
    \caption{$K=16$}
  \end{subfigure}  
  \begin{subfigure}[b]{0.23\textwidth}
    \includegraphics[width=\textwidth]{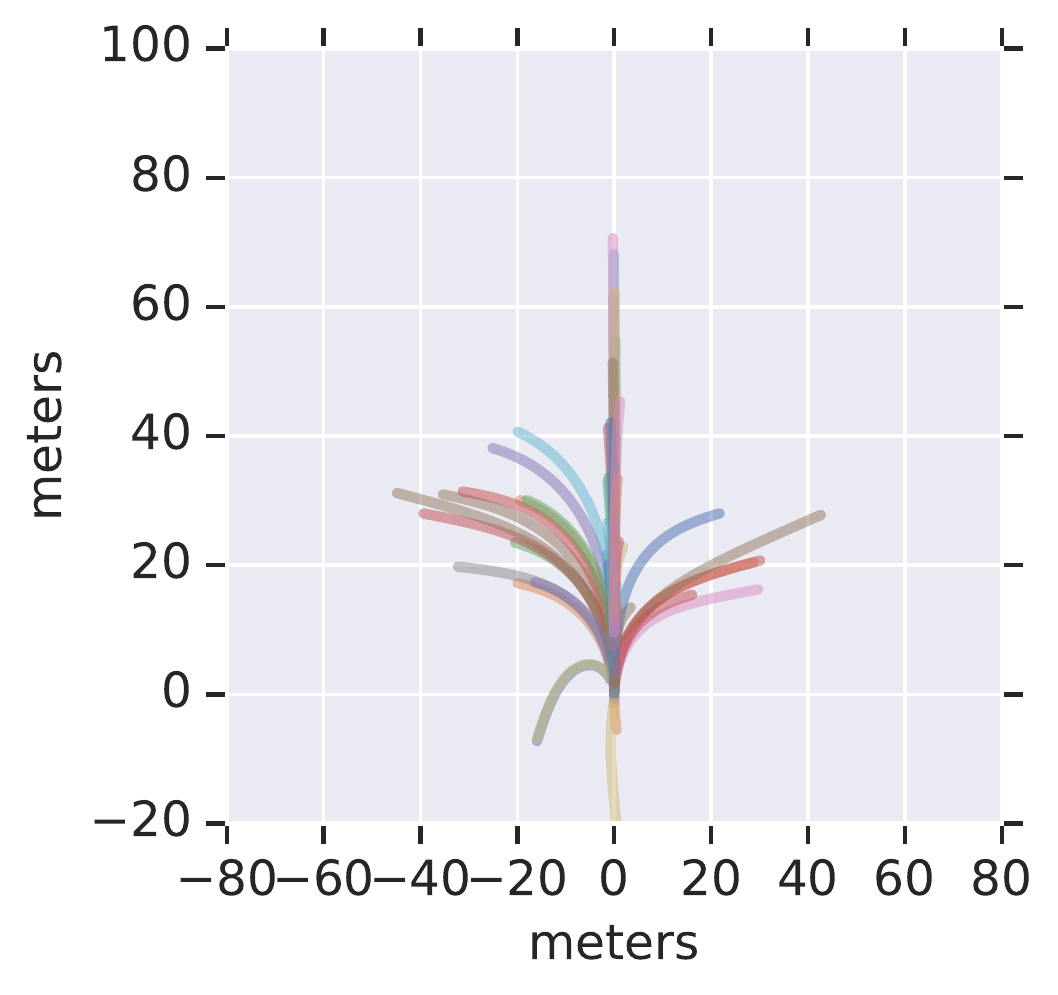}
    \caption{$K=32$}
  \end{subfigure}  
  \begin{subfigure}[b]{0.23\textwidth}
    \includegraphics[width=\textwidth]{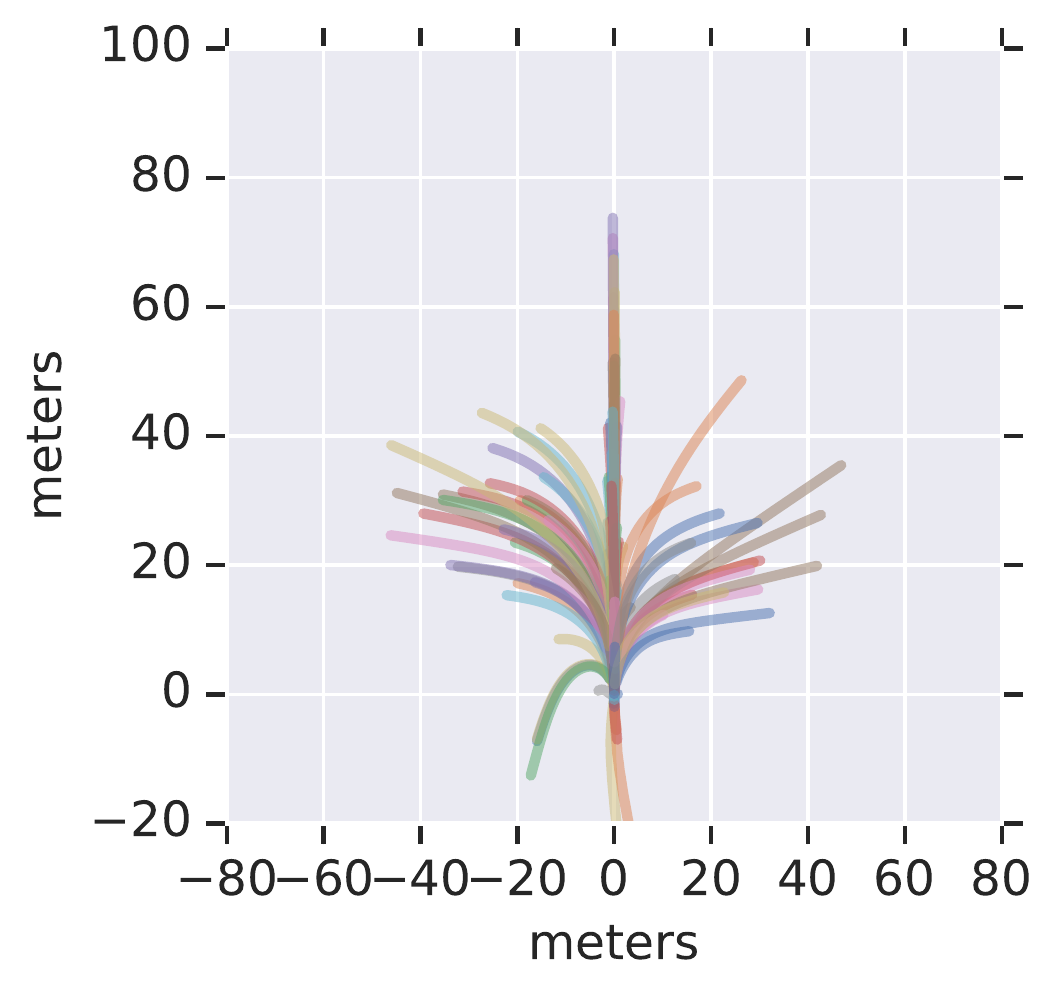}
    \caption{$K=64$}
  \end{subfigure}  
  \begin{subfigure}[b]{0.23\textwidth}
    \includegraphics[width=\textwidth]{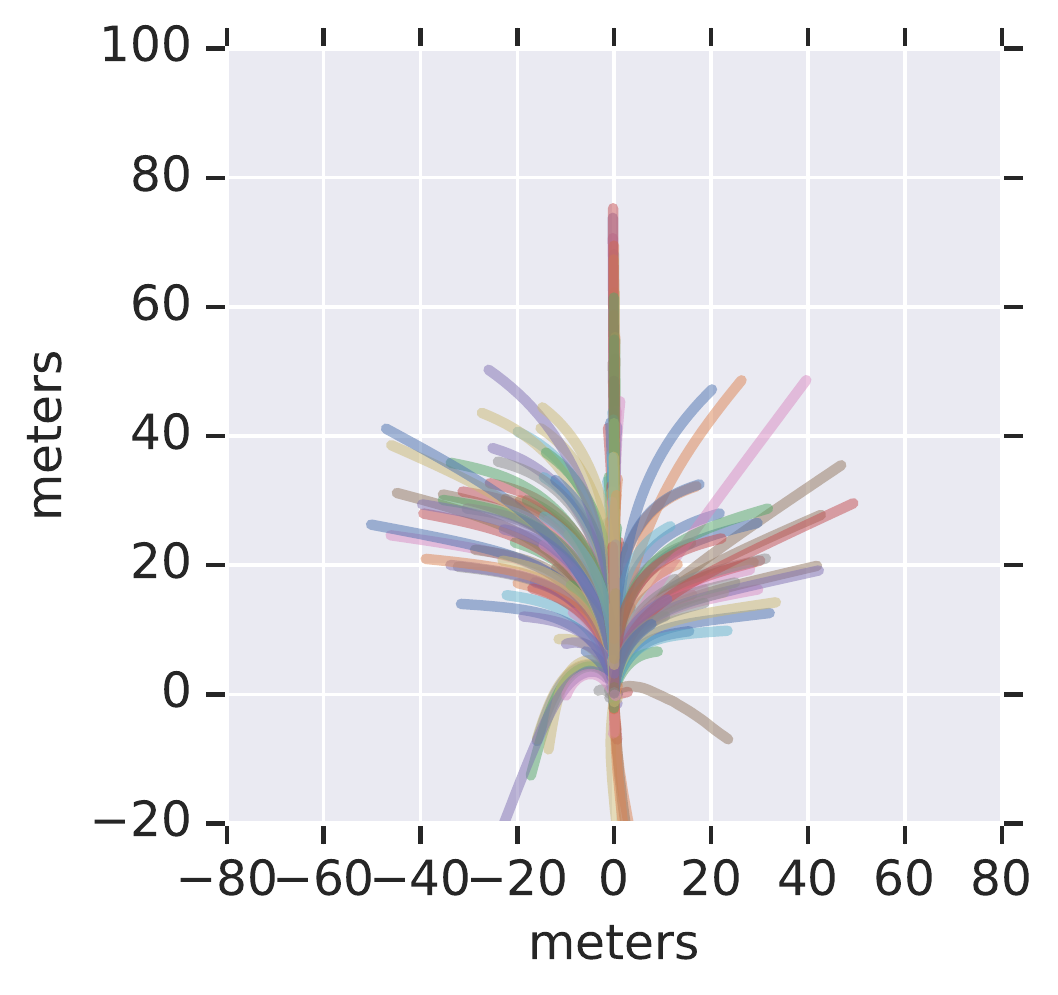}
    \caption{$K=128$}
  \end{subfigure}  
\caption{Visualization of trajectories and their clusters. \textbf{(a)}: 20k randomly sampled trajectories. \textbf{(b) - (g)}: Anchors for a variety of $K$ values.}
\label{fig:clusters_viz}
\end{figure*}

The anchors for the autonomous driving dataset for the various number of anchors $K$ are shown in \figref{fig:clusters_viz}. Since the data is captured in North America, left turns are expected to have a larger curvature than right turns, and u-turns mostly are on the left side. 

\section{Stanford Drone Dataset experimental details}
\label{sec:sdd_experiment_setup}

Unlike much of the past literature, we do not encode the raw location information directly.  In keeping with our top-down rendered representation as detailed in \secref{sec:method}, we render a history of oriented boxes as input channels concatenated with the current RGB video frame. Oriented boxes are estimated via past positions plus fixed-size extents.  For consistent processing, we resize and pad videos to have a resolution of 800 px $\times$ 800 px, maintaining the original aspect ratio, but report results at the original resolution.

Due to the small size of the dataset and lack of heading information, we created our anchor set for this dataset by enumeration rather than K-means: we chose 64 straight-trajectory anchors via 16 evenly-spaced orientations at 4 different final-waypoint distances (roughly 5\%, 10\%, 15\% and 20\% of the max image dimension), and 1 stationary anchor. Due to the complexity of natural image (RGB) input, we chose a higher-capacity feature processing backbone than other datasets: 28-layer ResNet, each layer of depth 32.  This proved beneficial over our default choice in Sec.~\ref{sec:bp}, and we did no other architecture search.

\section{CARLA experimental details}
\label{sec:carla_experiment_setup}

This dataset contains $60,701$ training, $7,586$ validation and $7,567$ test scenes from Town01 and $16,960$ test scenes from Town02. Each scene contains the autopilot's LIDAR observation together with 2 seconds of past and 4 seconds of future position information for the autopilot and 4 other agents at 5Hz. To allow comparison to the numbers reported in \cite{Rhinehart19}, we follow their training data setup: we use a $100\times100$ pixel crop of the top-down images, and only use the two LIDAR channels that represent a 2-bin histogram of points above and at ground-level in 0.5 m $\times$ 0.5 m cells.

\end{document}